\pdfoutput=1

\documentclass[11pt]{article}

\usepackage[preprint]{acl}
\usepackage{times}
\usepackage{latexsym}

\usepackage[T1]{fontenc}

\usepackage[utf8]{inputenc}

\usepackage{microtype}

\usepackage{inconsolata}

\usepackage{graphicx}
\usepackage{algorithmicx,algorithm}
\usepackage[noend]{algpseudocode}
\usepackage{algorithmicx,algorithm}
\usepackage{bm}
\usepackage{bbm}
\usepackage{booktabs}       
\usepackage{amsmath}
\usepackage{amsfonts}
\usepackage{array}
\usepackage{bm}
\usepackage{multirow}
\usepackage{subcaption}
\usepackage{hyperref}       
\usepackage{url}            
\usepackage{graphicx} 
\usepackage{utfsym}
\usepackage{xcolor}
\usepackage{inconsolata}
\usepackage{CJKutf8}
\usepackage{pythonhighlight} 
\usepackage{makecell}
\usepackage[normalem]{ulem}
\title{SpindleKV: A Novel KV Cache Reduction Method \\ Balancing Both Shallow and Deep Layers}

\author{
    Zicong Tang$^{2,\dagger}$, 
    Shi Luohe$^{2}$, 
    Zuchao Li$^{1,\dagger}$\thanks{$\ $  Corresponding author. $^\dag$ Equal contribution. This work was supported by the National Natural Science Foundation of China (No. 62306216), the National Social Science Fund of China (No. 24\&ZD186) and Xiaomi Open-Competition Research Program.}, \\
    \textbf{Baoyuan Qi}$^{3}$,
    \textbf{Guoming Liu}$^{3}$,
    \textbf{Lefei Zhang}$^{2}$, 
    \textbf{Ping Wang}$^{4}$ \\ 
    {$^{1}$School of Artificial Intelligence, Wuhan University} \\
    {$^{2}$School of Computer Science, Wuhan University} \\
    {$^{3}$Xiaomi, Beijing, China} \\
    {$^{4}$School of Information Management, Wuhan University} \\
    {\tt \{tangzc, shiluohe, zcli-charlie, zhanglefei, wangping\}@whu.edu.cn} \\
    {\tt \{qibaoyuan, liuguoming\}@xiaomi.com}\\
}

\begin{document}
\maketitle
\begin{abstract}
Large Language Models (LLMs) have achieved impressive accomplishments in recent years. However, the increasing memory consumption of KV cache has possessed a significant challenge to the inference system. 
Eviction methods have revealed the inherent redundancy within the KV cache, demonstrating its potential for reduction, particularly in deeper layers. However, KV cache reduction for shallower layers has been found to be insufficient. Based on our observation that, the KV cache exhibits a high degree of similarity.
Based on this observation, we proposed a novel KV cache reduction method, SpindleKV, which balances both shallow and deep layers. 
For deep layers, we employ an attention weight based eviction method, while for shallow layers, we apply a codebook based replacement approach which is learnt by similarity and merging policy.
Moreover, SpindleKV addressed the Grouped-Query Attention (GQA) dilemma faced by other attention based eviction methods.
Experiments on two common benchmarks with three different LLMs shown that SpindleKV obtained better KV cache reduction effect compared to baseline methods, while preserving similar or even better model performance.Our code is available in \url{https://github.com/tyxqc/SpindleKV}.
\end{abstract}

\section{Introduction}
Large language models (LLMs) ~\citep{OpenAITechnicalReport,llama2,BatGPTCharLie}, demonstrate impressive capabilities across various fields, such as machine translation~\citep{koshkin-etal-2024-transllama}, content generation~\citep{10.1145/3490099.3511105}, and harder tasks like coding~\citep{DBLP:journals/corr/abs-2308-12950} and reasoning~\citep{DBLP:conf/nips/Wei0SBIXCLZ22, yao-etal-2024-got}. LLMs' excellency in generating coherent text makes them useful tools in a wide range of industries. However, their more widespread and comprehensive use is facing a severe and realistic challenge, which is their high demand for GPU memory~\citep{YaoYao}. When using large models for inference, the memory overhead primarily consists of two parts: the model parameters and the context. The context exists in the form of Key-Value cache (KV cache), in order to reduce redundant computations during the autoregressive decoding~\citep{OPT,llama2,KCDLuohe,ModelHemorrhage,SirLLM,anonymous2025kvlatent}. 

With an increase in context length, the memory footprint for the KV cache increases proportionally. Occasionally, it can exceed the memory occupied by the model's parameters. As the context of LLMs getting longer, KV cache has become a new bottleneck that limits the deployment and application of LLMs~\citep{LongBench}. 

\begin{figure}
    \centering
    \includegraphics[width=\linewidth]{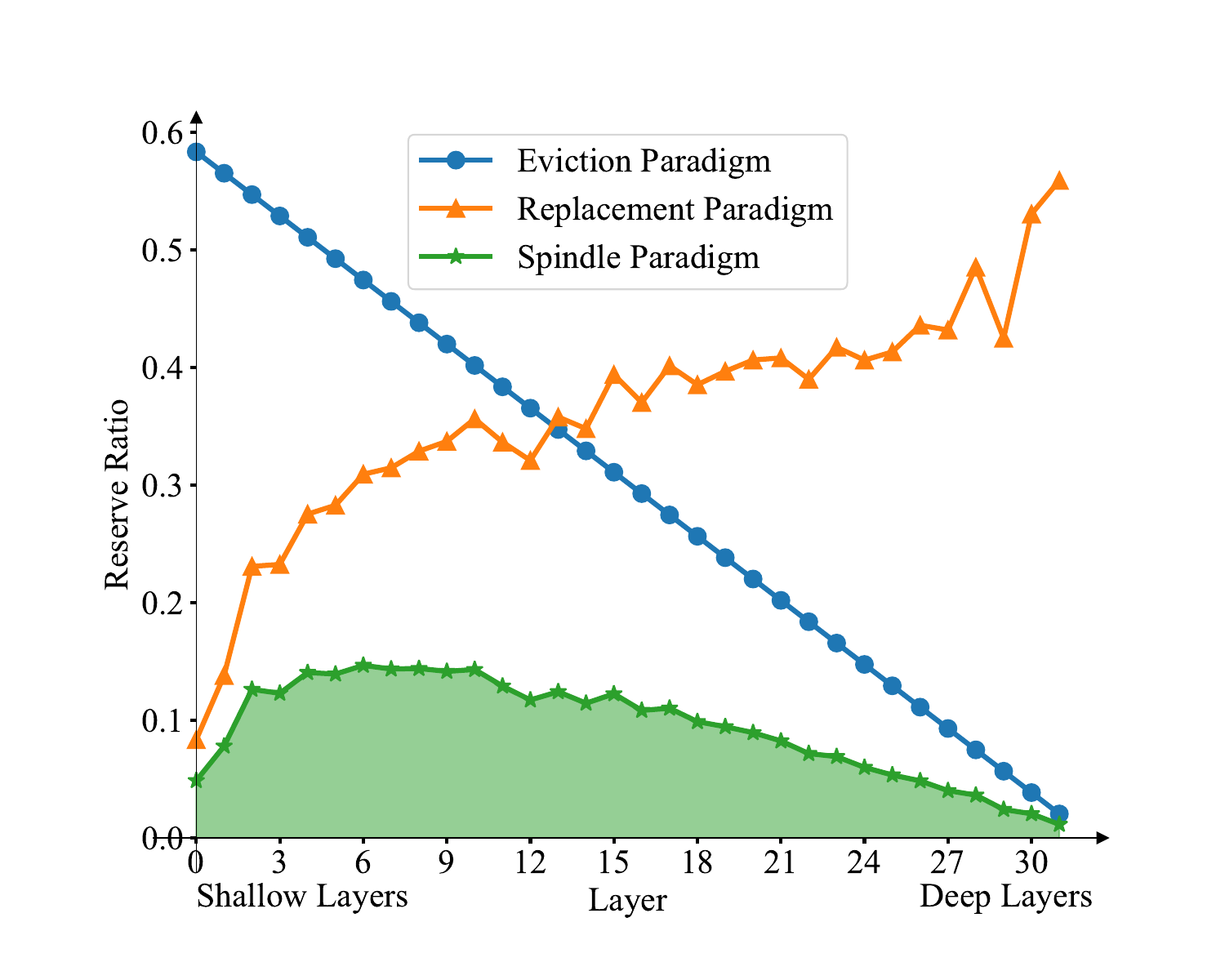}
    \caption{A quick illustration for how SpindleKV works on LLaMA-3-8b-instruct.}
    \label{fig:SpindleShape}
\end{figure}

Previous research has identified significant redundancy within KV cache, facilitating potential cache compression.
Token eviction~\citep{H2O,StreamingLLM,SnapKV,PyramidInfer,PyramidKV} is one of the prevailing methods. It reduces the size of the cache by eliminating tokens with smaller contribution in the attention mechanism.
Token merging~\citep{CaM,D2O,ModelTellsMerge} goes a step further by merging tokens based on their relevance, thereby compressing the information at a finer granularity. Another approach is quantization~\citep{ZipCache,MiniCache,AsymKV}, which provides a low-precision approximation of the KV cache without removing any information.
However, these works indicate that the reduction effectiveness is generally better for deeper layers than for shallower layers with token eviction, merging or quantization. Consequently, research on KV cache reduction methods for shallow layers has been largely overlooked in previous studies. 

KV cache reduction methods work better in deeper layers for two key reasons: (1) attention patterns in deeper layers naturally concentrate on fewer tokens, exhibiting low-rank characteristics; (2) deeper layers are more robust to modifications due to the shallow-to-deep propagation of changes through the network.
We argue that shallow layers also exhibit notable redundancy, primarily because tokens in these layers undergo fewer Transformer encoding iterations and thus receive limited contextual influence.
This indicates the existence of an additional form of redundancy beyond inter-token redundancy: tokens can be decomposed into smaller, redundant basis vectors that form their fundamental constituents.
By leveraging both inter-token redundancy in deeper layers and inner-token compositional redundancy in shallow layers, we propose SpindleKV to balance both deep and shallow layers KV cache reduction as illustrated in Figure~\ref{fig:SpindleShape}. Specifically, we employ token eviction mechanisms to eliminate redundancy in deeper layers, while adopting a Just-in-Time (JIT) learned basis vector codebook to reduce redundancy in shallow layers.

We conducted comprehensive evaluations across multiple models and datasets. The experimental results demonstrate that our approach achieves higher KV cache reduction rates while maintaining comparable performance, and delivers superior performance at equivalent reduction rates, compared to existing state-of-the-art (SOTA) methods.
Notably, in some settings, our approach enables a 50\% reduction in KV cache size without compromising model performance metrics.
Evaluations on long-text knowledge-intensive datasets demonstrate that SpindleKV preserves the model's knowledge retention and reasoning capabilities, even with significant KV cache reduction.
Further evaluations on the {\it Needle-in-a-Haystack} task confirm that SpindleKV maintains the model's long-sequence retrieval efficacy, which is essential for efficient large-scale document processing. Under identical KV cache compression ratios, SpindleKV demonstrates enhanced retrieval recall metrics, outperforming both PyramidInfer and PyramidKV.
Lastly, the comparative experiments conducted on the use of GQA-based models indicate that SpindleKV has superior compression capability on the GQA model, thereby validating that our approach is highly compatible with the new paradigms of LLMs.

\begin{figure*}[ht] 
\centering 
\includegraphics[width=0.95\linewidth]{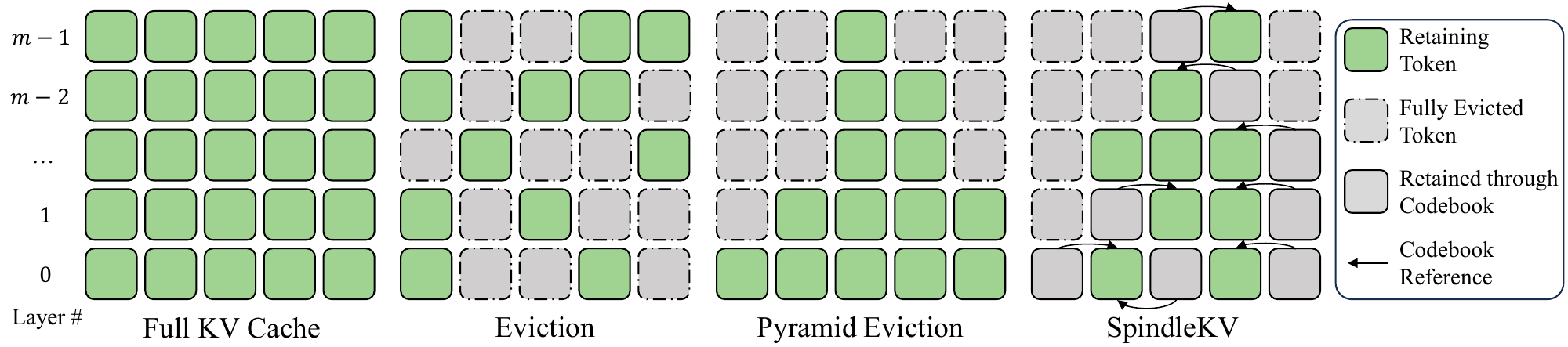} 
\caption{Comparison of major eviction methods.} 
\label{fig:FourCompressionMethod} 
\end{figure*} 

\section{Related Works}

\subsection{KV Head Reuse}
The concept of KV head reuse has been introduced in Multi-Query Attention (MQA,~\citealp{MQA}), where a single set of KV heads is retained to serve all Q heads. Even when sharing the same KV, different Q heads focus on different aspects of K, resulting in diverse combinations. The subsequent Grouped-Query Attention (GQA,~\citealp{GQA}) method introduced a new balance by grouping Q heads and sharing a single KV head within each group, offering a finer-grained approach to balancing performance and efficiency. However, despite GQA becoming the gold standard for LLMs, further compression of the KV cache presents challenges: (1) KV heads now possess higher information density; (2) KV head reuse restricts decisions to be made across all heads within the same group, limiting the potential for fine-grained methods.


\subsection{Token Eviction}
Earlier works discovered that tokens at the beginning and end of the sequence tend to be the most important in KV cache. These findings inspired several KV cache eviction methods like {\it StreamingLLM}~\citep{StreamingLLM}. Recent work has introduced eviction methods based on the contribution of attention scores, such as {\it H2O}~\citep{H2O} and {\it SnapKV}~\citep{SnapKV}. Building on this, some approaches no longer simply remove unimportant tokens, but instead merge them into the tokens that are retained, represented by {\it CaM}~\citep{CaM}, {\it D2O}~\citep{D2O}, and {\it KVMerger}~\citep{ModelTellsMerge}. Latest researches, {\it PyramidInfer}~\citep{PyramidInfer} and {\it PyramidKV}~\citep{PyramidKV}, show that the cost of evicting tokens from deeper layers is often lower, and the evicted KV cache exhibits a triangular pattern. However, these methods are difficult to integrate with GQA, as they require evaluating token acceptance or eviction for an entire group of Q heads, rather than for each individual Q head.

Quantization is also an important method to reduce KV cache though less relative with our method, detailed in Appendix~\ref{sec:quantization}.

In summary, existing KV cache reduction methods face challenges in compressing shallower layers. Moreover, attention score-based eviction methods exhibit poor compatibility with GQA.

\def \R{\mathbb{R}}
\def \rm#1{\mathrm{#1}}
\def \CS{\rm{cos\_sim}}

\begin{figure*}[ht]
    \centering
    \begin{subfigure}[b]{0.48\linewidth}
        \centering
        \includegraphics[width=\linewidth]{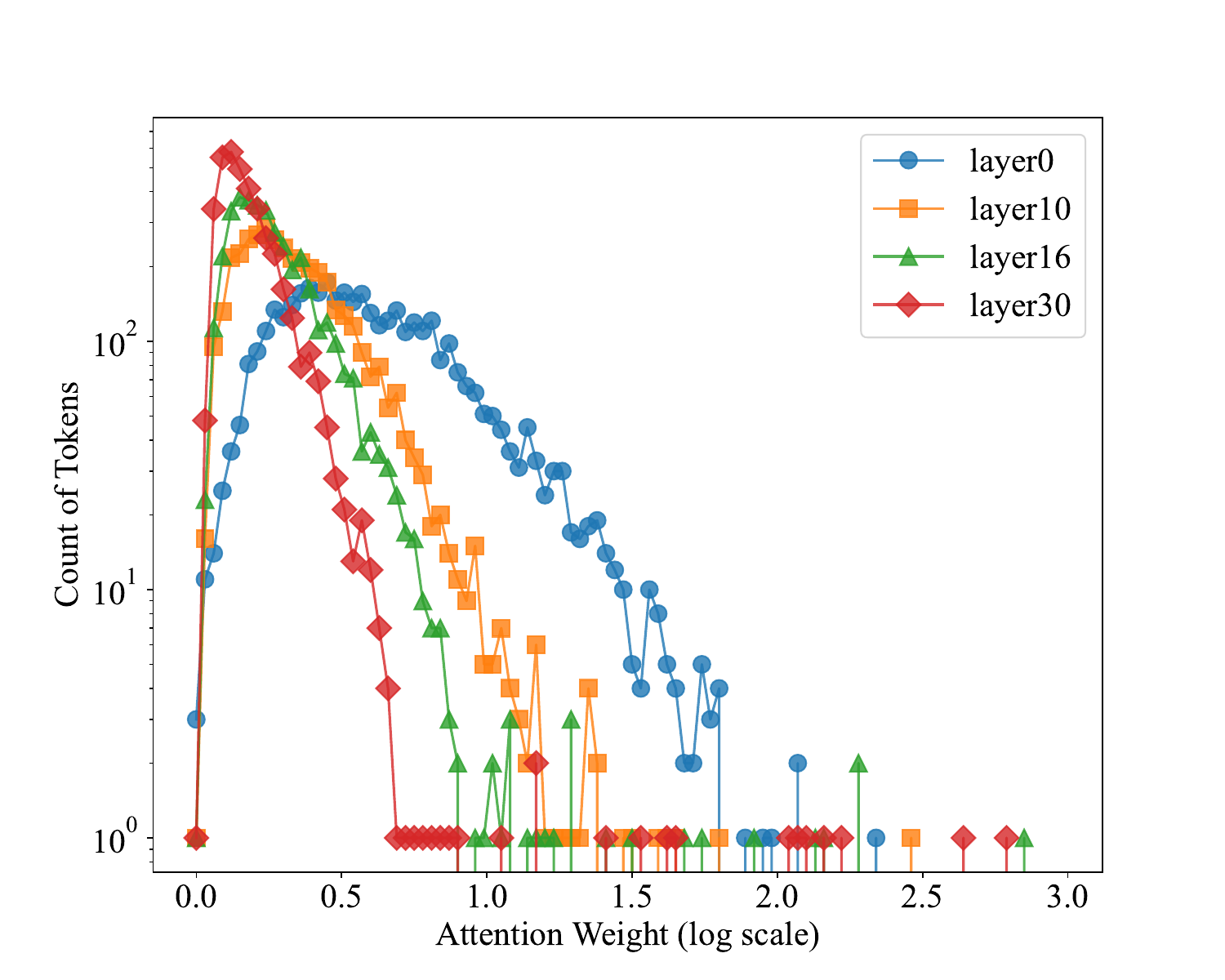}
        \caption{Sparse attention weight in deeper layers }
        \label{fig:AttentionWeight}
    \end{subfigure}
    \hfill
    \begin{subfigure}[b]{0.48\linewidth}
        \centering
        \includegraphics[width=\linewidth]{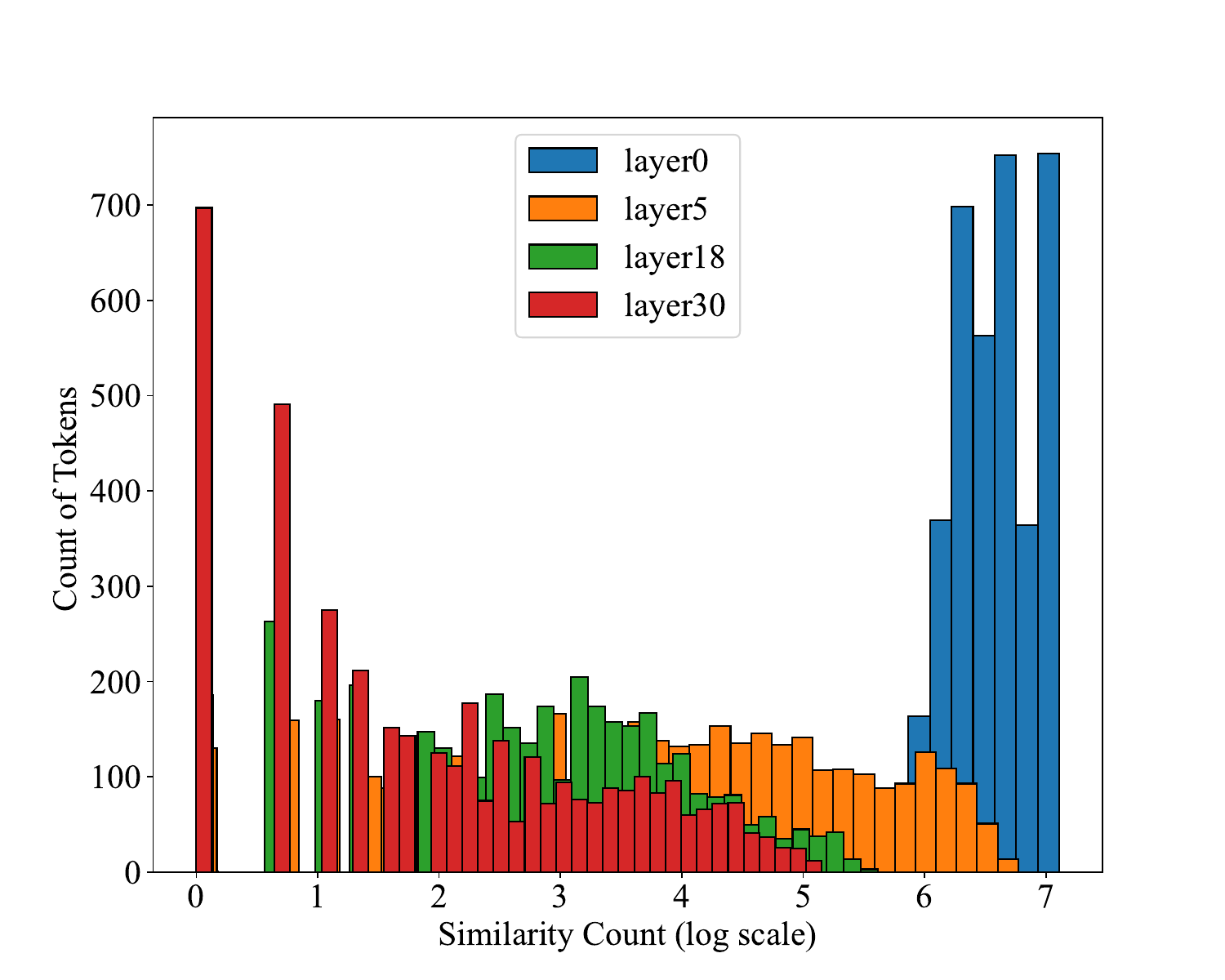}
        \caption{High cosine similarity in shallower layers}
        \label{fig:CosineSimilarity}
    \end{subfigure}
    \caption{The distribution of attention weight and cosine similarity in token level cross different layers of LLaMA2-7b-chat with a prompt sampling from 2WikiMQA.}
    \label{fig:Observation}
\end{figure*}

\section{Method}
\subsection{Notations}

SpindleKV does not involve information interaction between layers; therefore, we focus solely on the attention component within a single decoder layer. Let $d$ represent the hidden dimension of the model, $d_h$ the size of each attention head, and $h$ the total number of attention heads, $h_g$ as the number of KV heads for GQA models, and the amount of $Q$ heads per KV head $h_n=h/h_g$ correspondingly. We denote $l$ as the length of the input sequence $X\in \R^{l\times d}$. We firstly acquire the attention score $A_i$ of the $i$-th head through the equation:
\begin{equation}
\label{eq:attention}
\begin{aligned}
    A_i & = \rm{softmax}\left(\rm{mask}\left(
        \frac{Q_i \cdot K_i^{\top}}{\sqrt{d_h}}
    \right)\right) \\
    Q_i &= X \cdot W_{Q,i} \cdot \mathcal{R},
    \ K_i = X \cdot W_{K,i} \cdot \mathcal{R}
\end{aligned},
\end{equation}
in which $W_{\{Q,K\},i}$ denotes the parameters of the LLM and $\mathcal{R}$ as rotary position embedding matrix.
Then, the output of the masked multi-head self-attention is given out as follows.
\begin{equation*}
\begin{aligned}
    O &= \sum_{i=0}^{h-1} A_i \cdot V_i \cdot W_{O,i} \\
    V_i &= X \cdot W_{V,i}
\end{aligned}
\end{equation*}
Additionally, $W_{\{V,O\},i}$ denotes the other two parameter matrices. We define $\Gamma \in \{K, V\}$ as a unified representation for $K$ and $V$. Lastly we use $\{x, k_i, v_i, \gamma\}_{a}$ to represent the $a$-th entry of $\{X, K_i, V_i, \Gamma\}$. 

$A_i\in \R^{l\times l}$ is the critical standard for evaluating the contribution of each token to the attention. We typically use the accumulated attention scores, which represent the average attention score in a window $l_w$ for each query from a given token's key, as an indicator of the importance of a token. The accumulated attention score $ac_{i,a}$ for the $a$-th token in the sequence is calculated as shown in equation:
\begin{equation}
\label{eq:accatt}
    ac_{i,a} = \frac{\sum_{b=l-l_w}^{l-1} A_{i,a,b}}{l-a}.
\end{equation}
Additionally, for models that utilize GQA, the same $K$ would serves multiple $Q$. We've also have to take the average across all $h_n$ heads~\citep{PyramidInfer}, as equation:
\begin{equation}
    ac'_{i,a} = \frac{1}{h_n}\sum_{g=0}^{h_n}\frac{\sum_{b=l-l_w}^{l-1} A_{h_n\cdot i + g,a,b}}{n-a}.
\end{equation}

For SpindleKV, we also focused on the similarity $S_{\{K,V\}}$ within the KV cache, measured in {\it cosine similarity} as equation:
\begin{equation}
\begin{split}
    \CS\left(\gamma_a, \gamma_b\right) = \frac{a \cdot b^\top}{|a|\cdot |b|} \\
    S_{\Gamma,i,a,b} = \CS(\gamma_{i,a}, \gamma_{i,b}).
\end{split}
\end{equation}

\subsection{Preliminary Experiment}

We investigated the distributional characteristics of the KV cache components. To this end, we examined the attention distributions and the similarity of KV caches in the LLaMA2-7B-chat model on the 2WikiMQA dataset. 

\paragraph{Attention Sparsity}
We calculated the accumulated attention scores $ac$ for each token and each attention head across all layers, and then aggregated these scores in ascending order. As shown in Figure~\ref{fig:AttentionWeight}, we observed that as the layer depth increases, the number of tokens with lower attention scores also increases, reflecting the concentration of attention scores on a few specific tokens. This phenomenon reveals the attention sparsity in deeper layers and provides evidence for the feasibility of token eviction in these layers.

\paragraph{KV Constitutional Similarity}
We measured the number of token pairs in different layers whose similarity $S$ exceeds a threshold $\theta$. 
The composition of the KV cache exhibits a high degree of similarity in the shallower layers, as depicted in Figure~\ref{fig:CosineSimilarity}. This suggests that, in the shallower layers, although many tokens receive high attention scores, their constituent components are highly similar. More details are avaiable in Appendix~\ref{sec:MoreObservationResults}. This observation leads us to consider the potential of managing large numbers of similar KV caches using a codebook-based approach.


\subsection{Attention Weight Based Token Eviction}
\label{seq:Eviction}
We apply token eviction majorly for reducing the redundancy in deeper layers. 
Following PyramidKV~\citep{PyramidKV}, a simple linear interpolation works on the layer-wise KV cache allocation. We define $r$ as the total reserve ratio the KV cache. Following SnapKV~\citep{SnapKV}, we apply an observation window with length $l_w$ to calculate $ac$, in which all the tokens are reserved. The reserve ratio of the context, whose length $l_c = l-l_w$, is consistent across all decoding request.
We first calculate the reserve ratio of context $r_c$ by equation:
\begin{equation*}
    r_c = \frac{r\cdot l-l_w}{l_c}.
\end{equation*}
Moreover, we define the minimal preserve ratio for any layer as $\beta$. And we define $\alpha = \frac{1}{2}(1 + \beta)$.
Then we can define the maximal and minimal retain ratio for context for a model with $m$ layers, used by layer $0$ and $m-1$, in the equation:
\begin{equation}
\begin{aligned}
    r_c(0) &\! =\! \left\{
    \begin{aligned}
        & 2 \times r_c - 0.05, \!\!\!& \beta < r_c \leq \alpha \\
        & 1, & \alpha < r_c \leq 1
    \end{aligned}, 
    \right. \\
    r_c(m-1) &\! =\! \left\{
    \begin{aligned}
        & 0.05, &\ \beta < r_c \leq \alpha \\
        & 1 - 2 \times rc,& \alpha < r_c \leq 1
    \end{aligned} .
    \right. \\
\end{aligned}
\end{equation}
Finally, the retain ratio of $\lambda$-th layer $r_c(\lambda)$ can be determined by equation:
\begin{equation}
    r_c(\lambda) =r_c(0) + \frac{r_c(m-1) - r_c(0)}{m-1} \cdot \lambda.
\end{equation}

\begin{figure}[htbp]
    \centering
    \includegraphics[width=\linewidth]{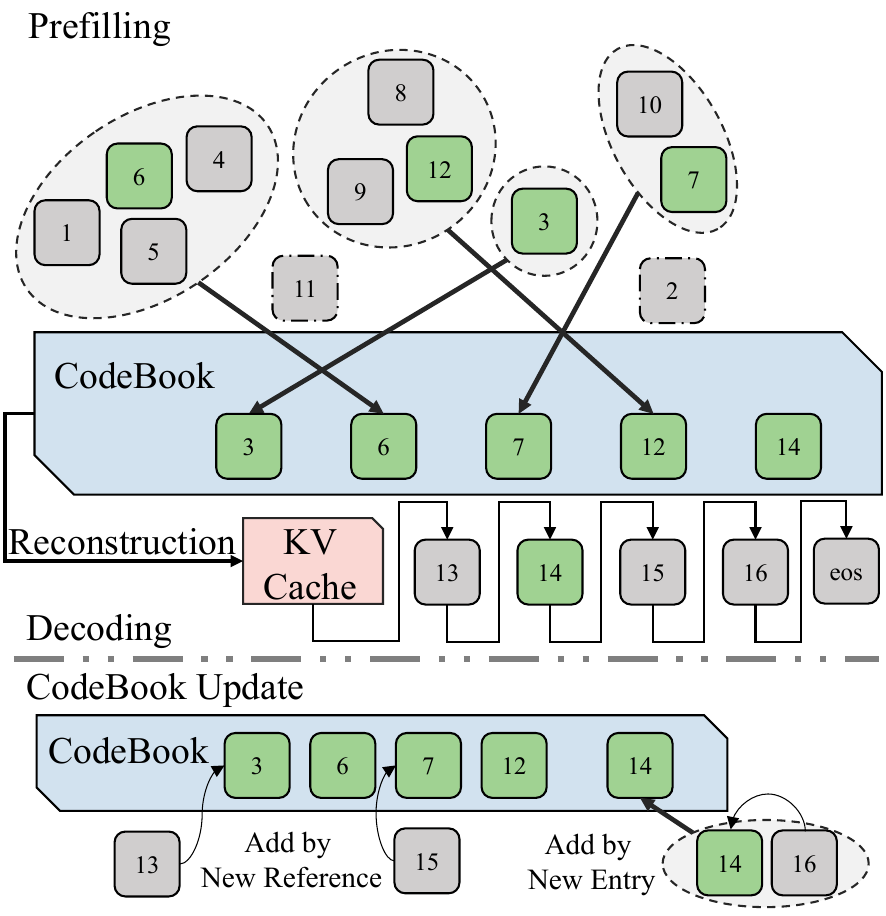}
    \caption{An overview of SpindleKV}
    \label{fig:CodeBook}
\end{figure}

At inference time, we dynamically select the preserved KV cache of $i$-th head by $ac_i$:
\begin{equation}
\label{eq:eviction}
\begin{aligned}
    \eta_i = \rm{argTop}&K\left(ac_i,\ k=\left\lfloor r_c(\lambda) \times l_c\right\rfloor \right) \\
    & \Gamma_{r,i} = \Gamma_i[\eta_i]
\end{aligned}.
\end{equation}

For models utilizing GQA, we can employ the $ac'$ designed for GQA in the previous section to complete this step. However, an alternative approach is to directly repeat these KV vectors $h_n$ times, effectively fully unfolding the GQA, then decision can then be made regarding whether to retain them. This step will increase the size of the KV vectors by several times, but we will address this overhead in the next subsection, ensuring that it does not require any additional storage space.



\begin{figure*}[t]
    \centering
    \begin{subfigure}[b]{0.30\linewidth}
        \centering
        \includegraphics[width=\linewidth]{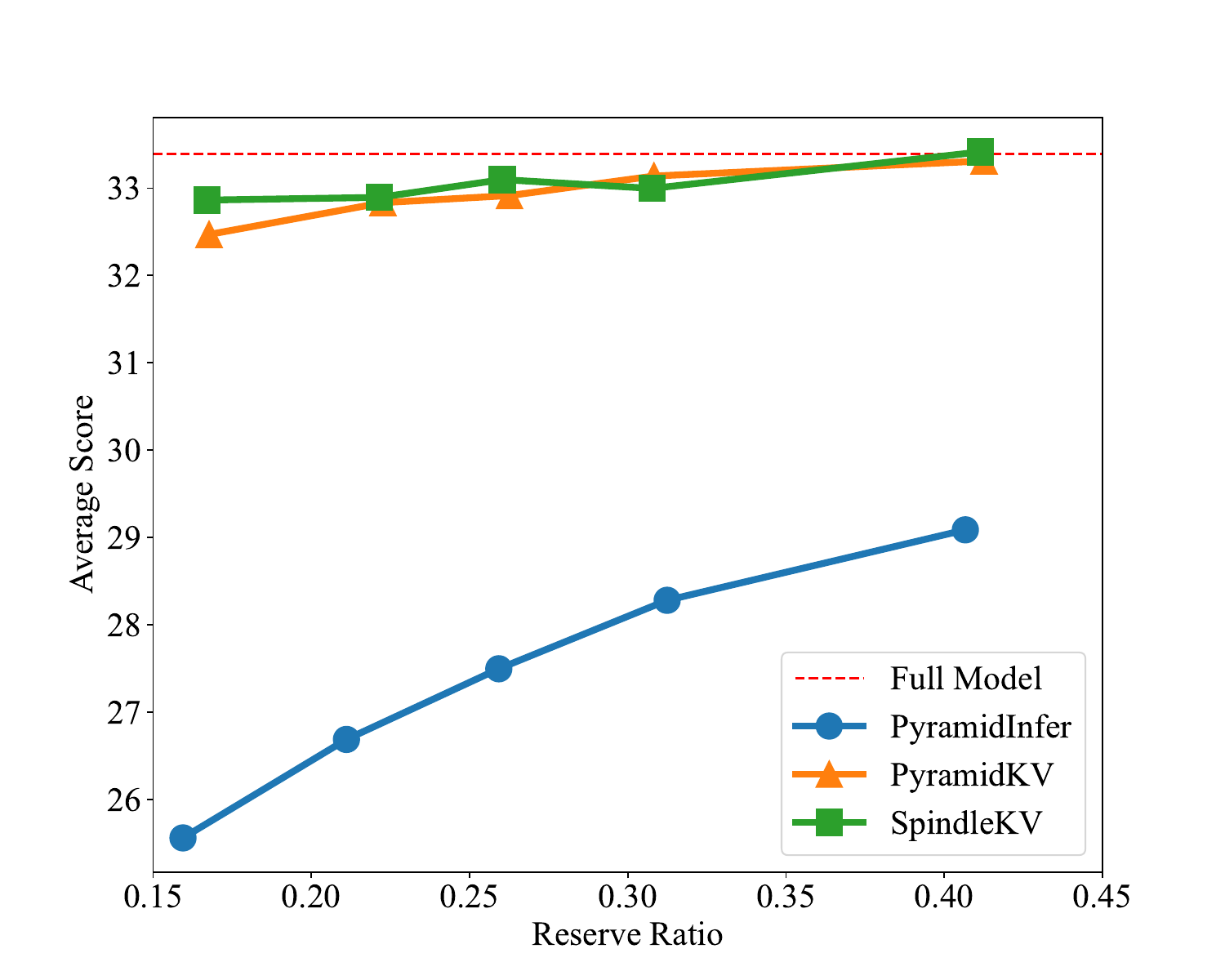}
        \caption{LongBench result of LLaMA2-7b}
        \label{fig:Llama2LongBench}
    \end{subfigure}
    \hfill
    \centering
    \begin{subfigure}[b]{0.30\linewidth}
        \centering
        \includegraphics[width=\linewidth]{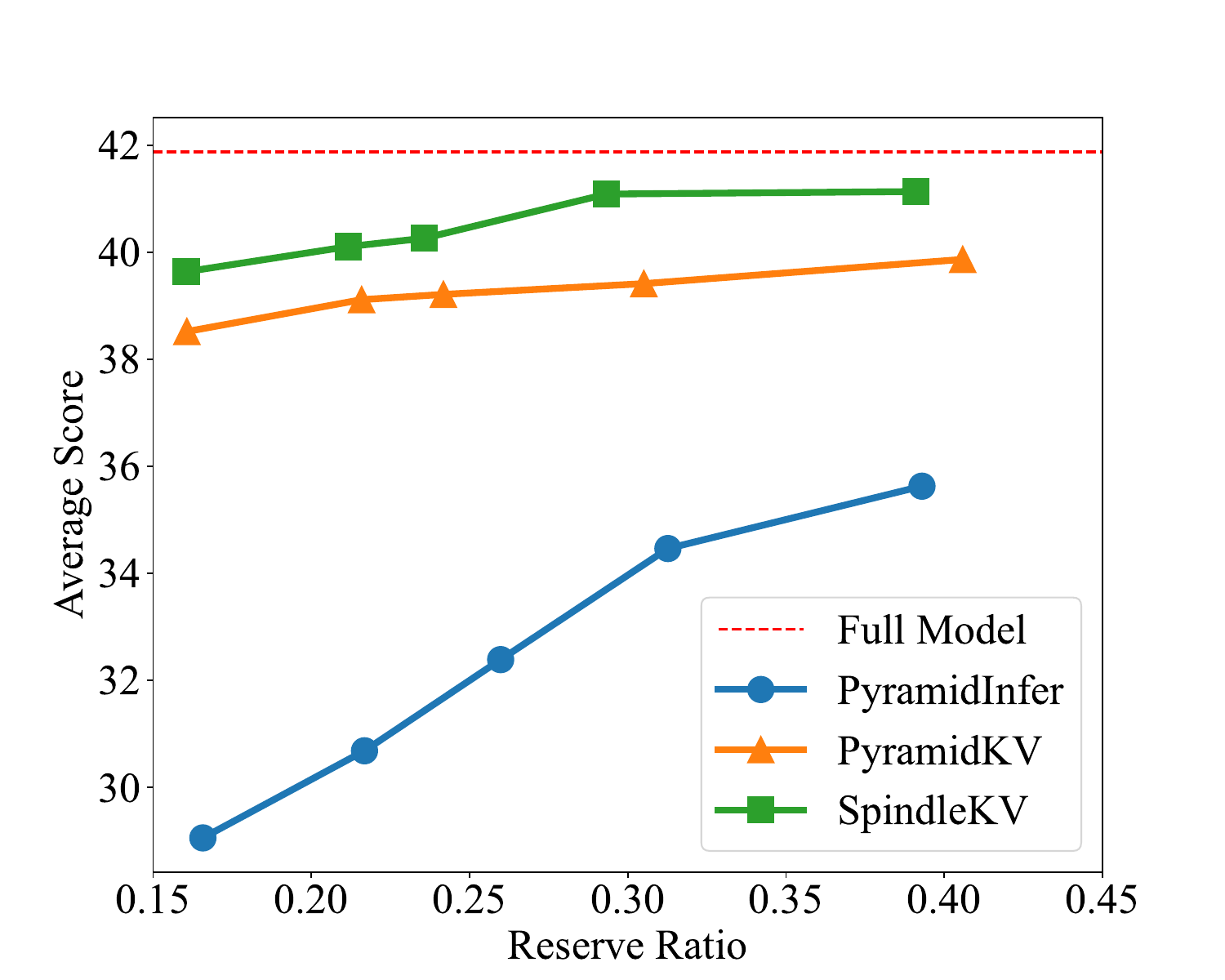}
        \caption{LongBench results of LLaMA3-8b }
        \label{fig:Llama3LongBench}
    \end{subfigure}
    \hfill
    \begin{subfigure}[b]{0.30\linewidth}
        \centering
        \includegraphics[width=\linewidth]{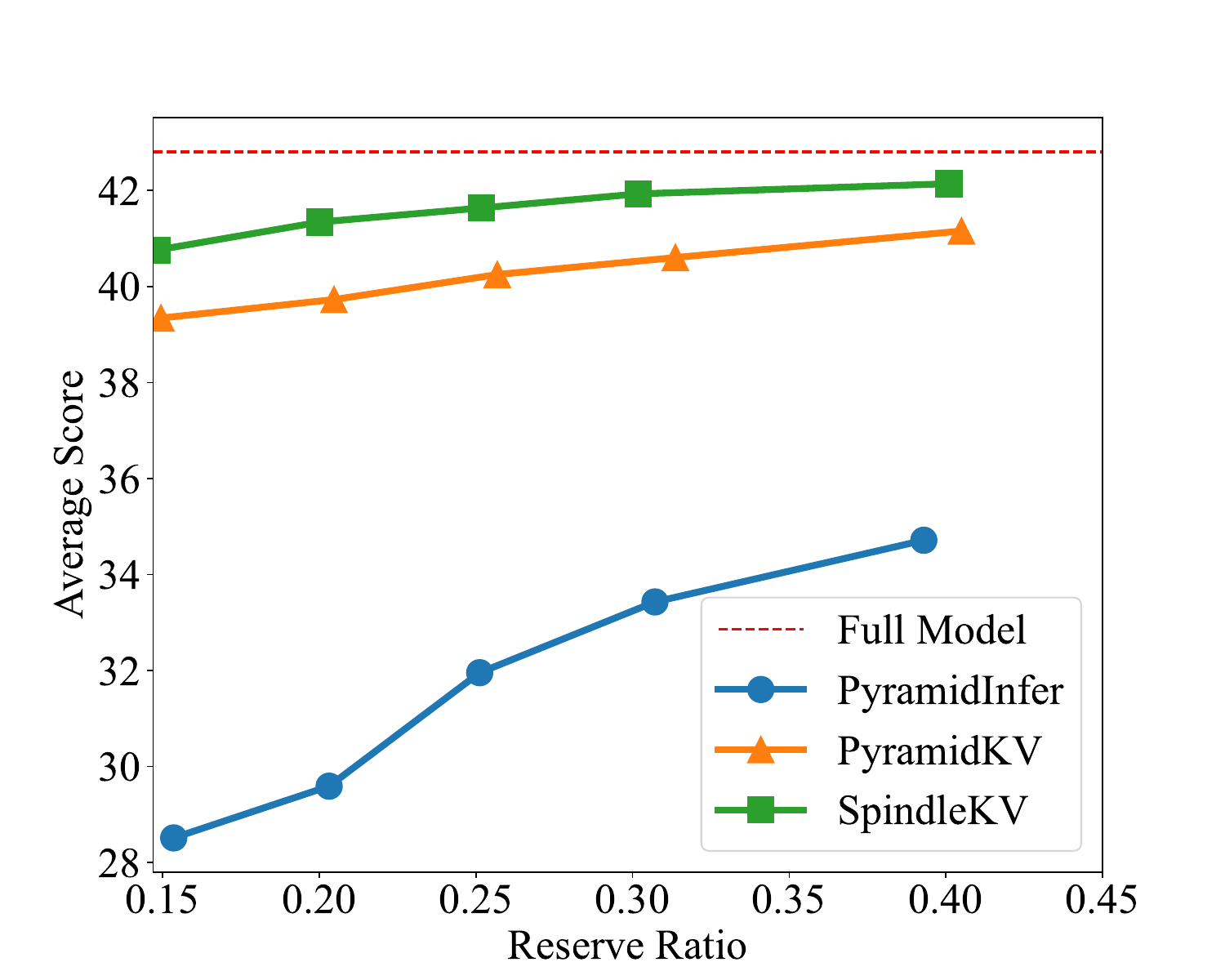}
        \caption{LongBench results of Mistral-7b}
        \label{fig:Mistral2LongBench}
    \end{subfigure}
    \caption{LongBench result on three models cross different reserve ratios.}
    \label{fig:LongBenchResult}
\end{figure*}

\subsection{Similarity Based Token Replacement}
\label{seq:Replacement}

The redundancy that the Eviction method cannot address arises from the high similarity between the constituent of the KV cache in shallower layers. Meanwhile, the operation of unfolding GQA in the previous section also increases this redundancy, as the cosine similarity of the unfolded components is clearly $1$. Therefore, we can approximate the KV cache by constructing a CodeBook from the similarity of the KV cache and retaining only the indices pointing to the CodeBook entries, thereby reducing its size.


In the prefilling stage, our goal is to create two CodeBooks $C_{\{K,V\}}$ that satisfies the following conditions:
\begin{itemize}
\item $\min(|C_K \cup C_V|)$
\item $\forall \gamma \in \Gamma_{r,i},\ \exists j \in \left[0, |C_\Gamma|-1\right], \\ 
\rm{s.t.}\ \CS(k, C_{\Gamma,j}) > \theta_\Gamma$,
\end{itemize}
where $\theta_{\{K,V\}}$ are the similarity threshold for $K$ and $V$.

However, we noted that cosine similarity only measures the direction of a vector and does not effectively capture its magnitude. This implies that we must also record the magnitude to minimize information loss as much as possible. Formally, this involves recording the magnitude $m_{\{K,V\}}$ of each KV cache, as shown in the equation:
\begin{equation}
\label{eq:1step}
    m_\Gamma  = \sqrt{\Gamma \cdot \Gamma^\top},\ \Gamma_r  \longleftarrow \frac{\Gamma_r}{m_\Gamma}.
\end{equation}

Then we construct a binary matrix $G_{\{K,V\}}$, where $G_{\Gamma,a,b}$ indicates whether the $a$-th and $b$-th tokens can be merged, determined by:
\begin{equation}
\label{eq:2step}
    G_\Gamma = \rm{where}(S_\Gamma > \theta_\Gamma,\ 1,\ 0).
\end{equation}

This matrix also represents the adjacency matrix of an undirected graph. What we need to identify is the node with the highest degree in this graph, as the vector corresponding to this node can represent the greatest number of other vectors. The degree of each node is computed as shown in the equation:
\begin{equation}
\label{eq:r1step}
    s_{\Gamma, a}=\sum_{b=0}^{N-1}G_{\Gamma,a,b}.
\end{equation}
We greedily choose the token with the highest degree and add it to the $C$. Then we delete this vertex and the vertices that are directly adjacent to this vertex from the graph through a simple mask operation. Formally, each round of the CodeBook building process is given by:
\begin{equation}
\label{eq:r2step}
\begin{aligned}
    \iota &= \rm{argmax}(s_\Gamma) \\
    C_\Gamma &\longleftarrow C_\Gamma + [\Gamma_{r,\iota}] \\
\end{aligned},
\end{equation}
then, we build the reference $r_{\{K,V\}}$ of each token to the corresponding vector in the CodeBook:
\begin{equation}
\label{eq:r3step}
\begin{aligned}
    \eta_{\iota} &= \rm{argwhere}(G_{\Gamma, \iota}==1) \\
    &r_\Gamma[\eta_\iota] = |C_\Gamma| - 1 \\
\end{aligned}.
\end{equation}
The $\rm{mask}_{\{K, V\}}$ and the clean up to the graph is given by:
\begin{equation}
\label{eq:r4step}
\begin{aligned}
    &\rm{mask}_\Gamma = \lnot G_{\Gamma, \iota}^\top \cdot \lnot G_{\Gamma, \iota} \\
    &G_\Gamma \longleftarrow G_\Gamma \ \&\  \rm{mask}_\Gamma
\end{aligned}.
\end{equation}
We firstly execute equation~\ref{eq:1step},~\ref{eq:2step}, then execute equation~\ref{eq:r1step},~\ref{eq:r2step},~\ref{eq:r3step},~\ref{eq:r4step} repetitively until $G_{\Gamma}==\mathbf{0}$. Then we finished the construction of CodeBook $C_\Gamma$, magnitudes $m_\Gamma$ and the reference to CodeBook for every token $r_\Gamma$. The process is illustrated in Algorithm~\ref{al:TokenReplacement}.

\begin{figure}[t]
    \centering
    \begin{algorithm}[H]
        \caption{CodeBook Generate of $\Gamma \in \{K, V\}$}
        \label{al:TokenReplacement}
        \hspace*{0.02in} {\bf Input:} Cached $\Gamma_r$, Threshold $\theta_\Gamma$. \\
        \hspace*{0.02in} {\bf Output:} CodeBook $C_\Gamma$, References $r_\Gamma$, \\ Magnitudes $m_\Gamma$
        \begin{algorithmic}
            \State $C_\Gamma \gets \emptyset$ 
            \State $r_\Gamma \gets [-1, -1, \dots, -1]$ \Comment{Initialization}
            \State $m_\Gamma \gets L_2{\it Norm}(\Gamma, \rm{dim}=-1)$
            \State $\Gamma_r \gets \Gamma_r / m_\Gamma$ \Comment{Normalize $\Gamma$ to 1}
            \State $S_\Gamma \gets \CS(\Gamma,\ \Gamma)$
            \State $G_\Gamma \gets \rm{where}(S_\Gamma > \theta_\Gamma, 1, 0)$
            \While {$G_\Gamma != \mathbf{0}$}
                \State $s_\Gamma \gets \rm{sum}(S_\Gamma,\rm{dim}=1)$
                \State $\iota \gets \rm{argmax}(s_\Gamma)$
                \State $C_\Gamma \gets C_\Gamma + [\Gamma_{r,\iota}]$ \Comment{New entry}
                \State $\eta_\iota \gets \rm{argwhere}(G_{\Gamma, \iota} == 1)$
                \State $r_\Gamma[\eta_\iota] \gets |C_\Gamma| - 1$ \Comment{Inserted to the end}
                \State $\rm{mask}_\Gamma \gets \rm{matmul}(\lnot G_{\Gamma, \iota}^\top,\lnot G_{\Gamma, \iota})$
                \State $G_\Gamma \gets G_\Gamma\ \&\ \rm{mask}_\Gamma$ \Comment{Clean up}
            \EndWhile
            \Return $C_\Gamma$, $r_\Gamma$, $m_\Gamma$
        \end{algorithmic}
    \end{algorithm}
\end{figure}

At inference time, we can reconstruct KV cache efficiently through the equation:
\begin{equation}
    \Gamma_r = C_\Gamma[r_\Gamma] \otimes m_\Gamma
\end{equation}

During inference, we generate a large number of new KV cache entries. We first search for suitable CodeBook entries for merging, using $\theta$ as the threshold. If such entries are found, we merge them; otherwise, we repeat the process of building the CodeBook for the remaining tokens.

It is important to note that although this search process has a slightly higher time complexity, it does not result in a significant additional time overhead. Furthermore, our method operates on the pre-RoPE $K$, meaning that after reconstruction, the RoPE operation must be re-applied. However, due to limitations imposed by memory bandwidth and the inherent sparsity of RoPE as a sparse matrix multiplication, this step does not introduce substantial time overhead~\citep{KIVI}. This is because, while the computational workload is increased, the corresponding increase in arithmetic intensity enhances the peak FLOPS achievable by the chip~\citep{10.1145/1498765.1498785}.

\begin{table*}[t]
\centering
\renewcommand{\arraystretch}{1.6} %
\resizebox{\textwidth}{!}{
\begin{tabular}{lcccccccccccccccccc}
\toprule
\multirow{2}{*}{\textbf{Methods}}& \multirow{2}{*}{\textbf{Ratio}} & \multicolumn{3}{c}{\textbf{Single-Document QA}} & \multicolumn{3}{c}{\textbf{Multi-Document QA}} & \multicolumn{3}{c}{\textbf{Summarization}} & \multicolumn{3}{c}{\textbf{Few-shot Learning}} & \multicolumn{2}{c}{\textbf{Synthetic}} & \multicolumn{2}{c}{\textbf{Code}} & \multirow{2}{*}{\textbf{AVG.}} \\
\cmidrule(lr){3-5} \cmidrule(lr){6-8} \cmidrule(lr){9-11} \cmidrule(lr){12-14} \cmidrule(lr){15-16} \cmidrule(rl){17-18}
 && Na.QA & Qasp & Mu.QA & Ho.QA & Wi.QA & Musq & Gv.Rp & QMSm & M.New & TREC & Tr.QA & SASm & PCnt & Pa.Rt & Lcc & RB.P & \\
\midrule
FullKV  &100\% & 25.70 & 29.75 & 41.12 & 45.55 & 35.87 & 22.35 & 25.63 & 23.03 & 26.21 & 73.00 & 90.56 & 41.88 & 4.67 & 69.25 & 58.05 & 50.77 & 41.46 \\
\midrule
PyramidInfer &39.3\%  &23.75	&17.46	&29.97	&35.08	&23.92	&16.90	&28.08	&21.26	&24.42	&62.00	&85.06	&41.45	&1.04	&41.23	&50.95	&52.86	&34.71 \\
PyramidKV &40.5\% &26.31	&28.64	&\textbf{49.12}	&41.66	&25.98	&\textbf{19.02}	&26.38	&23.91	&22.66	&70.00	&85.88	&42.53	&2.69	&\textbf{86.32}	&54.04	&53.36	&41.16 \\
SpindleKV &40.1\% &\textbf{26.95}    &\textbf{31.23}	&49.01	&\textbf{41.89}	&\textbf{26.90}	&18.60	&\textbf{29.92}	&\textbf{24.56}	&\textbf{24.89}	&\textbf{71.50}	&\textbf{85.98}	&\textbf{43.39}	&\textbf{2.74}	&86.18	&\textbf{56.39}	&\textbf{54.13}	&\textbf{42.14} \\
\midrule
PyramidInfer &30.7\% &22.65	&14.57	&30.14	&33.87	&23.52	&15.59	&26.82	&21.10	&23.10	&61.00	&84.18	&40.57	&1.85	&32.25	&50.64	&53.02	&33.43 \\
PyramidKV &31.4\%  &26.12	&27.31	&48.31	&41.44	&25.14	&\textbf{18.72}	&25.28	&23.61	&21.99	&69.00	&86.27	&42.65	&2.53	&84.81	&53.71	&52.77	&40.60 \\
SpindleKV &30.2\% &\textbf{26.69}	&\textbf{30.19}	&\textbf{49.32}	&\textbf{42.09}	&\textbf{27.23}	&18.69	&\textbf{28.52}	&\textbf{24.19}	&\textbf{24.02}	&\textbf{71.00}	&\textbf{86.38}	&\textbf{43.54}	&\textbf{3.03}	&\textbf{87.18}	&\textbf{55.19}	&\textbf{53.62}	&\textbf{41.93}\\
\midrule
PyramidInfer &25.1\% &21.43	&13.49	&25.88	&31.92	&20.44	&14.83	&25.27	&20.57	&22.06	&58.00	&82.16	&40.77	&1.45	&27.93	&52.53	&52.55	&32.00\\
PyramidKV &25.7\% &25.14	&26.11	&46.97	&40.56	&25.46	&\textbf{19.06}	&25.00	&23.32	&21.55	&\textbf{70.50}	&\textbf{86.41}	&41.92	&\textbf{3.26}	&84.56	&51.95	&52.23	&40.25\\
SpindleKV &25.2\% &\textbf{25.97}	&\textbf{30.34}	&\textbf{49.17}	&\textbf{42.06}	&\textbf{27.35}	&18.22	&\textbf{27.99}	&\textbf{24.25}	&\textbf{23.19}	&70.00	&86.22	&\textbf{42.96}	&2.74	&\textbf{87.26}	&\textbf{54.69}	&\textbf{53.77}	&\textbf{41.64} \\
\midrule
PyramidInfer &20.3\% &18.73	&11.96	&24.34	&27.10	&15.87	&11.98	&24.89	&20.07	&21.19	&54.00	&76.19	&39.92	&2.06	&20.83	&51.83	&52.50	&29.59\\
PyramidKV &20.5\% &24.96	&25.19	&47.12	&39.94	&25.45	&\textbf{18.63}	&24.05	&23.35	&20.71	&69.50	&85.48	&41.87	&\textbf{2.48}	&83.56	&52.02	&51.35	&39.73\\
SpindleKV &20.0\% &\textbf{26.03}	&\textbf{29.49}	&\textbf{49.61}	&\textbf{41.87}	&\textbf{26.50}	&17.98	&\textbf{27.01}	&\textbf{23.73}	&\textbf{22.73}	&\textbf{70.00}	&\textbf{86.28}	&\textbf{43.59}	&2.29	&\textbf{86.99}	&\textbf{53.81}	&\textbf{53.57}	&\textbf{41.34}\\
\midrule
PyramidInfer &15.4\% &17.28	&11.52	&23.41	&26.38	&16.72	&11.98	&24.49	&19.41	&20.77	&53.00	&68.87	&39.78	&\textbf{3.53}	&11.88	&\textbf{54.48}	&\textbf{52.72}	&28.51\\
PyramidKV &15.0\% &24.36	&24.66	&45.85	&40.67	&24.81	&17.83	&23.29	&23.41	&20.53	&\textbf{70.00}	&86.09	&40.74	&3.27	&81.84	&51.54	&50.60	&39.34\\
SpindleKV &14.8\% &\textbf{26.02}	&\textbf{28.05}	&\textbf{48.84}	&\textbf{40.74}	&\textbf{25.45}	&\textbf{19.05}	&\textbf{25.51}	&\textbf{23.72}	&\textbf{21.95}	&69.50	&\textbf{86.13	}&\textbf{42.47}	&2.70	&\textbf{85.65}	&53.89	&52.48	&\textbf{40.76}\\
\bottomrule
\end{tabular}
}
\caption{LongBench results for Mistral-7b-instruct-v0.2. The correspondence between abbreviations and datasets is provided in Appendix~\ref{seq:DatasetMapping}. We ensure that the reserve ratio $r$ of SpindleKV is slightly lower than the baselines, with the deviation kept within 2\%, as our method cannot precisely control the KV cache size.}
\label{tb:LongBenchResult}
\vspace{-3mm}
\end{table*}

\begin{table*}[htbp]
    \centering
    \renewcommand{\arraystretch}{1.2} %
    \scalebox{0.78}{
        \begin{tabular}{l|cccccccc}
            \toprule
            \textbf{Methods} &
            \textbf{Ratio} &
            \textbf{Na.QA} &
            \textbf{Wi.QA} &
            \textbf{QMSm} &
            \textbf{Tr.QA} &
            \textbf{PCnt} &
            \textbf{Lcc} &
            \textbf{AVG.} \\
            \midrule
            FullKV       &100\%  & 24.50 & 36.23 & 23.40 & 90.48 & 4.77 & 59.21 & 39.77 \\
            \midrule
            \textit{w/o.} repeat &39.1\% & \textbf{24.00} & 28.02 & 22.64 & 89.05 & 4.95 & 58.24 &37.82 \\
            \textit{w/.} repeat    &39.1\% & 23.87 & \textbf{36.02 }& \textbf{23.28} & \textbf{90.43} & \textbf{5.23} & \textbf{59.37} &\textbf{39.70} \\
            \midrule
            \textit{w/o.} repeat &30.4\% & 21.47 & 24.00 & 22.21 & 89.47 & 4.70 & 57.63 & 36.58 \\
            \textit{w/.} repeat    &29.3\% & \textbf{24.18} & \textbf{34.95} & \textbf{23.52} & \textbf{90.43} & \textbf{5.24} & \textbf{59.24} & \textbf{39.59} \\
            \midrule
            \textit{w/o.} repeat &21.3\% & 21.21 & \textbf{24.91} & 22.30 & 89.95 & 4.75 & 55.86 & 36.50 \\
            \textit{w/.} repeat    &21.2\% & \textbf{23.92} & 33.90 &\textbf{ 22.66} & \textbf{90.56} & \textbf{5.58} & \textbf{58.26} & \textbf{39.15} \\
            \bottomrule
        \end{tabular}   
    }
    \caption{Comparison between SpindleKV with and without repeating on LLaMA3-8b-instruct when handling GQA.}
    \label{tb:AnotherApproachofGQA}
\end{table*}


\section{Experiments}
\subsection{Experimental Setup}
We conducted our experiments on three models, including LLaMA2-7b-chat~\citep{llama2}, LLaMA3-8b-instruct~\citep{Llama3} and Mistral-7b-instruct-v0.2~\citep{mistral7b}. Their maximum context length ranges from 4k to 32k. LLaMA2-7b employs MHA, while LLaMA3-8b and Mistral-7b utilize GQA with $h_n=8$, $h_g=4$. All models' $h=32$.

We evaluate the model's knowledge and reasoning capabilities on {\it LongBench}~\citep{LongBench}, which contains 16 long-context knowledge intensive subsets covering 6 tasks including multi-document question answering (QA), single-document QA, summarization, few-show learning, synthetic tasks and code, with the length of most tasks ranging from $5$k to $15$k. We also evaluate our method on {\it Needle-in-a-Haystack} task~\citep{Needle} to test the long-context retrieval ability.

We select {\it Pyramidinfer}~\citep{PyramidInfer} and {\it PyramidKV}~\citep{PyramidKV} as baselines, both of which compress the KV cache based on token eviction method and allocate pyramid shaped KV cache cross layers.
\begin{equation}
\label{eq:metrics}
\begin{aligned}    
    & \left\{
    \begin{aligned}
        r_{1}^\lambda &= \frac{
            \sum_{j=0}^{h_{g}-1}(|K_\lambda^{j,r}|+|V_\lambda^{j,r}|)
        }{
            \sum_{j=0}^{h_{g}-1}(|K_j^i|+|V_j^i|)
        } \\
        r_{2}^\lambda &= \frac{
            |C^\lambda_{K} \cup C^\lambda_{V}|
        }{
            \sum_{j=0}^{h_{g}}(|K_{j,r}^\lambda|+|V_{j,r}^\lambda|)
        } \\
        r_{3}^\lambda &= \frac{1}{d_h}\left(
            d_h + \frac{\rm{int\_bit}}{\rm{key\_bit+1}}
        \right) \\
    \end{aligned}
    \right. \\
    & r^\lambda=r_1^\lambda \times r_2^\lambda \times r_3^\lambda \\
    & r = \frac{1}{m} \sum_{\lambda=0}^{m-1}r^\lambda \\
\end{aligned}
\end{equation}

We use the reserve ratio $r$ to measure the compression intensity. We define $r^\lambda$ as the reserve ratio for $\lambda$-th layer, which is co-determined by $r_{1}^\lambda$, the reserve ratio of eviction method, $r_{2}^\lambda$, the reserve ratio of replacement method, and $r_{i,3}$, serves as the \texttt{dtype} convert ratio
since we store {\tt int} type index and {\tt float} type magnitude for each key and value. The calculation is given by Formura~\ref{eq:metrics}. $r^\lambda=r_{1}^\lambda$ for pyramidinfer and PyramidKV, and $h_{g}=h$ if a repeat operation is conducted before eviction. We record $r_1$ and $r_2$ for every input when evaluating on {\it LongBench} to ensure their precision.

\begin{table}
\centering
\renewcommand{\arraystretch}{1} 
\begin{tabular}{cl}
\toprule
\textbf{Hyper-parameter} & \textbf{Value} \\
\midrule
Key Threshold (\(\theta_K\))   & 0.98 \\
Value Threshold (\(\theta_V\)) & 0.95 \\
$\beta$               & 0.05 \\
$\alpha$             & 0.525 \\
\bottomrule
\end{tabular}
\caption{Hyper-parameters Used in Experiments}
\label{tab:hyperparams_vertical}
\end{table}

\begin{table}
    \centering
    \renewcommand{\arraystretch}{1.2} %
    \scalebox{0.9}{
        \begin{tabular}{l|cc}
            \toprule
             \textbf{Methods}&
             \textbf{LLaMA3-8b}&
             \textbf{Mistral-7b} \\
             \midrule
             PyramidInfer
             &0.615 &0.621 \\
             PyramidKV &0.938 &0.962 \\
             SpindleKV &\textbf{0.979} &\textbf{0.975}\\
             \bottomrule
        \end{tabular}
    }
    \caption{Accuracy of {\it Needle-in-a-Haystack} on LLaMA3-8b-instruct and Mistral-7b-instruct-v0.2 with $15\%$ KV cache reserved.}
    \label{tb:Needle-in-a-Haystack result}
\end{table}

 \begin{figure*}[t]
    \centering
    \begin{subfigure}[t]{0.48\linewidth}
        \centering
        \includegraphics[width=\linewidth]{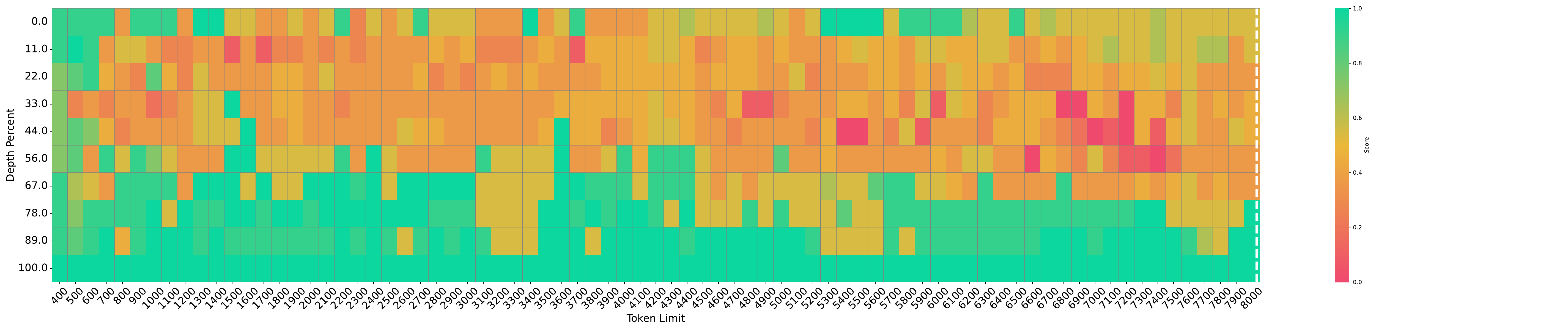}
        \caption{PyramidInfer on LLaMA3-8b.}
    \end{subfigure}%
    \hfill
    \begin{subfigure}[t]{0.48\linewidth}
        \centering
        \includegraphics[width=\linewidth]{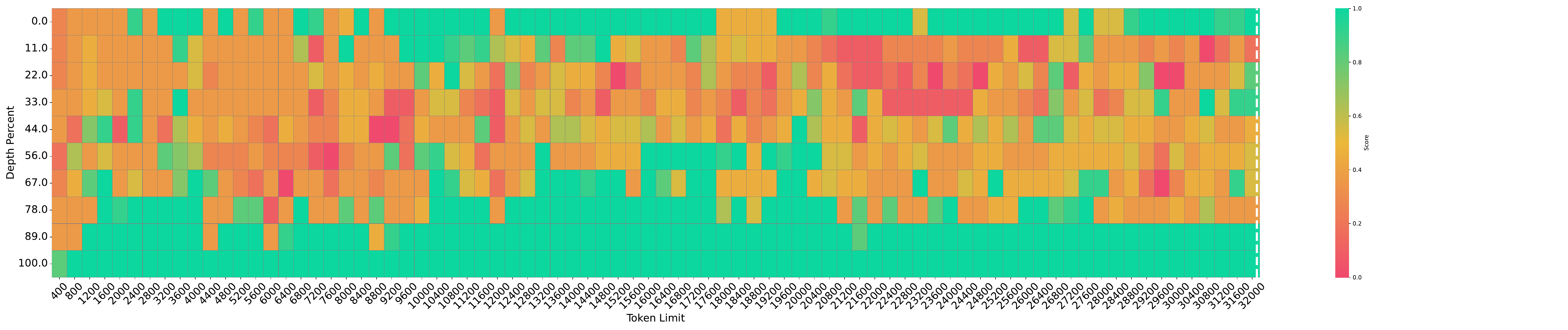}
        \caption{PyramidInfer on Mistral-7b.}
    \end{subfigure}

    \vspace{1em} 

    \begin{subfigure}[t]{0.48\linewidth}
        \centering
        \includegraphics[width=\linewidth]{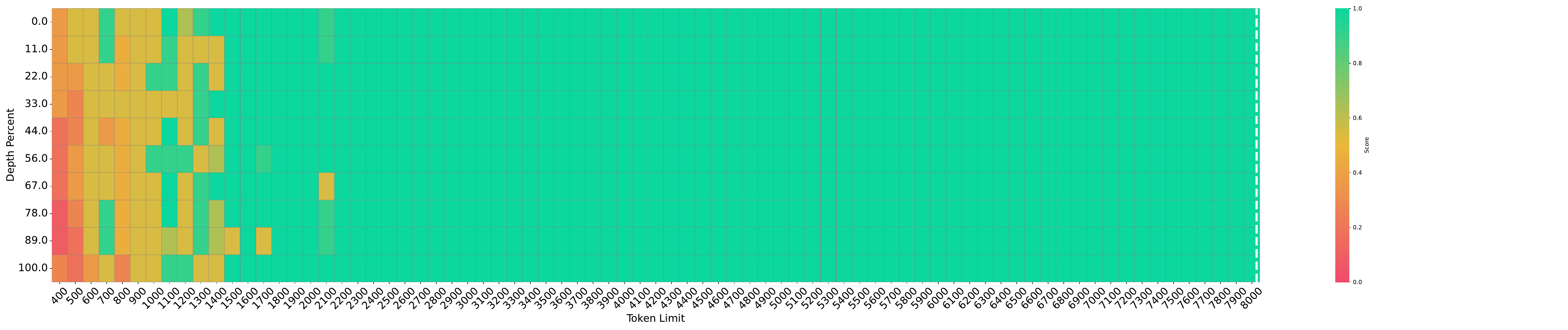}
        \caption{PyramidKV on LLaMA3-8b.}
    \end{subfigure}%
    \hfill
    \begin{subfigure}[t]{0.48\linewidth}
        \centering
        \includegraphics[width=\linewidth]{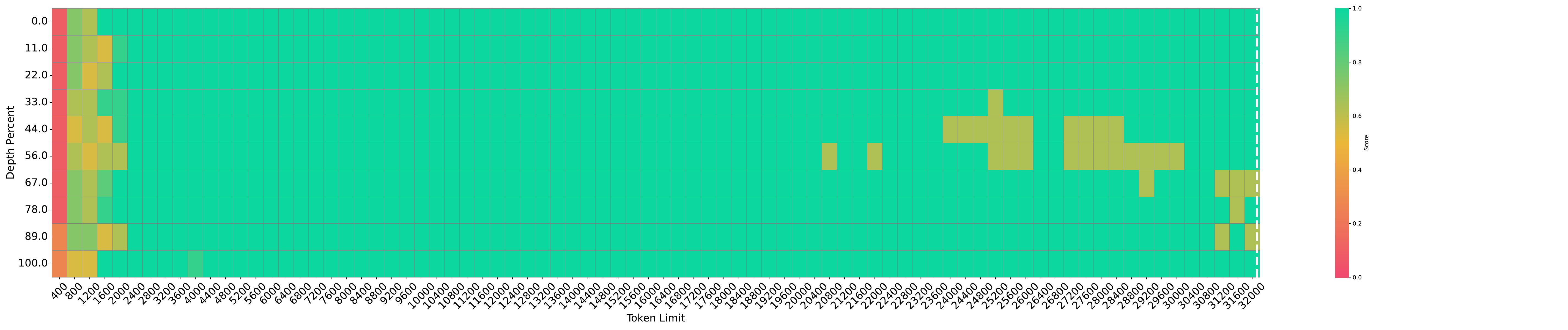}
        \caption{PyramidKV on Mistral-7b.}
    \end{subfigure}

    \vspace{1em} 

    \begin{subfigure}[t]{0.48\linewidth}
        \centering
        \includegraphics[width=\linewidth]{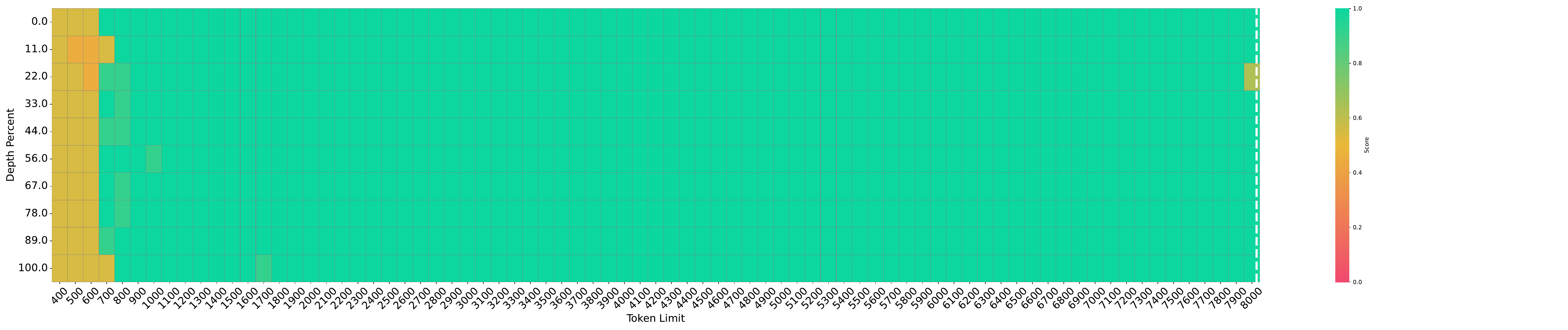}
        \caption{SpindleKV on LLaMA3-8b.}
    \end{subfigure}%
    \hfill
    \begin{subfigure}[t]{0.48\linewidth}
        \centering
        \includegraphics[width=\linewidth]{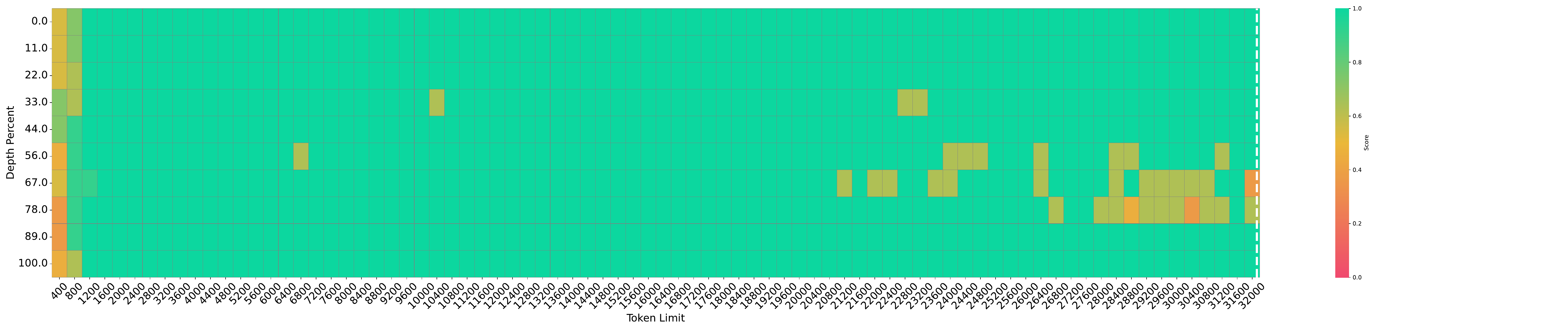}
        \caption{SpindleKV on Mistral-7b.}
    \end{subfigure}

    \caption{Visualization of Needle-in-a-Haystack. The vertical axis of the table
    represents the depth percentage, and the horizontal axis represents the token length.}
    \label{fig:VisualizationNeedle}
\end{figure*}

\begin{table*}[htbp]
    \centering
    \renewcommand{\arraystretch}{1.2} %
    \scalebox{0.9}{
        \begin{tabular}{c|ccccccc}
            \toprule
            \textbf{Ratio} &
            \textbf{Na.QA} &
            \textbf{Wi.QA} &
            \textbf{QMSm} &
            \textbf{Tr.QA} &
            \textbf{PCnt} &
            \textbf{Lcc} &
            \textbf{AVG.} \\
            \midrule
            100\% & 18.40 & 25.73 & 20.97 & 83.38 & 5.50 & 60.70 & 35.78 \\
            47.8\% & 18.57 & 25.51 & 21.05 & 84.40 & 6.00 & 60.54 & 36.01 \\
            29.4\% & 15.99 & 24.26 & 19.68 & 80.62 & 6.00 & 53.20 & 33.29 \\
            \bottomrule
        \end{tabular}
    }
    \caption{Accuracy of 6 datasets on SpindleKV without eviction on LLaMA2-7b-chat.}
    \label{tb:CodeBookOnly}
\end{table*}

\begin{table*}[htbp]
    \centering
    \scalebox{0.9}{
        \begin{tabular}{l|cccccccc}
        
            \toprule
            \textbf{Methods} &
            \textbf{Ratio} &
            \textbf{Na.QA} &
            \textbf{Wi.QA} &
            \textbf{QMSm} &
            \textbf{Tr.QA} &
            \textbf{PCnt} &
            \textbf{Lcc} &
            \textbf{AVG.} \\
            \midrule
            FullKV&100\% & 18.40 & 25.73 & 20.97 & 83.38 & 5.50 & 60.70 & 35.78 \\
            \midrule
            \textit{w/o.} reconstruct  &19.7\% & 17.17 & \textbf{25.64} & \textbf{20.73} & \textbf{84.15} & 5.50 & 59.34 & 35.42 \\
            \textit{w/.} reconstruct&20.5\% &\textbf{ 17.34} & \textbf{25.64} & 20.64 & \textbf{84.15}  & \textbf{6.00} & \textbf{60.31} & \textbf{35.68} \\
            \midrule
            \textit{w/o.} reconstruct &28.3\% & 18.06 & 25.70 & 20.76 & \textbf{84.04} & 5.50 & 59.06 & 35.52 \\
            \textit{w/.} reconstruct &29.1\% & \textbf{18.13} & \textbf{26.10} & \textbf{20.95} & 83.69 & \textbf{6.00} &\textbf{ 59.85} & \textbf{35.79} \\
            \midrule
            \textit{w/o.} reconstruct &39.4\% & \textbf{18.47} & \textbf{25.77} & \textbf{20.81} & 83.89 & 5.56 & 59.41 & 35.65 \\
            \textit{w/.} reconstruct&40.2\%& 18.45 & 25.74 & 20.66 & \textbf{84.31} & \textbf{6.00} & \textbf{60.45} & \textbf{35.94} \\
            \bottomrule
        \end{tabular}
    }
    \caption{Accuracy of 6 datasets on SpindleKV with and without reocnstruct opearation on LLaMA2-7b-chat.}
    \label{tb:MagnRecon}
\end{table*}

Hyper-parameters including $\alpha$, $\beta$ and $\theta$, are given in Table~\ref{tab:hyperparams_vertical}.


\subsection{Accuracy on Long Context Tasks}
\paragraph{LongBench Result} 
We evaluate SpindleKV on three models mentioned above and the result are illustrated in Figure~\ref{fig:LongBenchResult} and Table~\ref{tb:LongBenchResult}. Figure~\ref{fig:LongBenchResult} shows that our method can further compress KV cache while retaining model's capability. Our method outperforms the two baselines on both models with and without GQA, showing that our method effectively reduced the constituent redundancy in shallower layers. In models with GQA structure, we surpass baselines with only half of the KV cache compared with them, which indicates that SpindleKV has a better compatibility with GQA.  Table~\ref{tb:LongBenchResult} shows results of Mistral-7b on different datasets in {\it LongBench}, more results are avaiable in Appendix~\ref{sec:MoreLongBenchResults}. We report that our method further surpasses Pyramidinfer on average score and outperforms PyramidKV on 13 datasets between different reserve ratios on Mistral-7b. The results on {\it LongBench} show that our method can further reserve model's  knowledge retention and reasoning capabilities with the same KV cache budget.

\paragraph{Needle-in-a-Haystack Result}
Following PyramidKV, we use LLaMA3-8b-isntruct and Mistral-7b-instruct-v0.2 for this task. The results are displayed in Table~\ref{tb:Needle-in-a-Haystack result} and more details are available in Figure~\ref{fig:VisualizationNeedle}. In our experiment, models reserve only $15\%$ KV cache then retrieve a special "needle" from the context. The results in Table~\ref{tb:Needle-in-a-Haystack result} indicate that, compared with PyramidInfer and PyramidKV, SpindleKV significantly maintains the model's long-sequence retrieval efficacy with the same KV cache budget. As illustrated in Figure~\ref{fig:VisualizationNeedle}, PyramidInfer, which apply a mean operation on attention weight before eviction raise significantly context loss. And for PyramidKV, also raise noticeable context loss when reserve only $15\%$ KV cache. But SpindleKV saves more context with the same KV cache budget, significantly improved the quality of after-compress retrieval.

\subsection{Ablation Study}
We investigate the correctness of the settings and the effectiveness of every component in SpindleKV.
\paragraph{Integrating GQA}
We discussed two approaches of how eviction methods integrate with GQA in Section~\ref{seq:Eviction}. We conduct an experiment with LLaMA3-8b-instruct on the other approach following PyramidInfer, which applies the same averaged attention weight on different heads to conduct the eviction. The results in Table~\ref{tb:AnotherApproachofGQA} indicate that our method further outperforms this approach, significantly addressed the GQA dilemma.

\paragraph{CodeBook Works Without Eviction}
We remove the eviction part of SpindleKV and compress KV cache with only the cosine similarity based replacement method. We evaluate this method on 6 datasets of {\it LongBench} with $50\%$ and $30\%$ KV cache reserve ratio for LLaMA2-7b-chat. The results in Table~\ref{tb:CodeBookOnly} indicates that this method compresses half of KV cache without any impact on accuracy and reserves most of model's capabilities with $30\%$ KV cache, Which confirms that there is a significant amount of constituent redundancy in the KV cache, which appears in the form of Constitutional cosine similarity.

\paragraph{Effectiveness of reconstruction Via magnitude}
In SpindleKV, we record the magnitude for each key and value, and reconstruct them to their original magnitude after indexing them from the CodeBook. We compare the performance of SpindleKV with and without the reconstruction operation on LLaMA2-7b-chat. The results in Table~\ref{tb:MagnRecon} indicate reconstruct operation can effectively reserve model's capability with only a slight memory consumption of the magnitude.  

\section{Conclusion}
In this study, we find the constituent redundancy of KV cache in shallower layers and develop SpindleKV. It addresses the GQA dilemma faced by other attention weight based eviction methods and balances both shallow and deep layers with an attention weight based eviction method and a CodeBook based replacement approach. Experimental results present SpindleKV is a promising solution on long-context inference with constrained memory for KV cache.

\section*{Limitations}
In this work, we develop SpindleKV, which achieves a balance in KV cache compression across shallow and deep layers. Experiments conducted on two long-context benchmarks and three models demonstrate the effectiveness of our method.

While our current approach shows promising results, future work will focus on further refining the control over KV cache size to achieve more precise management. Additionally, although we have validated the effectiveness of our method on LLaMA2-7b-chat, LLaMA3-8b-instruct, and Mistral-7b-instruct-v0.2, we plan to extend our evaluation to additional models such as Qwen2.5-7b~\citep{Qwen2.5}, LLaMA2-13b, and LLaMA3-70b. This will allow us to further demonstrate the generality of our approach across a broader range of settings.

\bibliography{anthology, main}
\appendix
\clearpage
\section{Quantization}
\label{sec:quantization}
Quantization is the most straightforward method for providing a low-precision approximation of the KV cache. {\it FlexGen}~\citep{FlexGen} and {\it KIVI}~\citep{KIVI}, as prior works, have proved the capability of quantization. Indeed, as the bit-width decreases, uniform quantization of the entire KV cache quickly encounters limitations, prompting the development of more fine-grained approaches~\citep{MiKV}. One such approach, {\it AsymKV}~\citep{AsymKV}, identifies a similar pattern observed in eviction methods, namely, that quantization compression becomes easier for deeper layers of the cache. However, similarly, it has significant limitations when it comes to compressing the KV cache in shallower layers.

\section{More Observation Results}
\label{sec:MoreObservationResults}
We conducted observations of cosine similarity on Key and Values, and the results are showed in Figure~\ref{fig:CosineSimilarityKey} and Figure~\ref{fig:CosineSimilarityValue}. We set $\theta=0.9$ for Key and $\theta=0.6$ for Value as the token level cosine similarity is more obvious in Key. Different with the past study~\citep{ModelTellsMerge}, we also observed the layer level decrease of cosine similarity in Value. In the shallower layers, the highly similar constituent components is consistent in Key and Value. 




\section{Abbreviation-Dataset Mapping}
\label{seq:DatasetMapping}
Due to space constraints, we use abbreviations for datasets in Table~\ref{tb:LongBenchResult},~\ref{tab1:longbench2} and ~\ref{tab1:longbench3}. Below~\ref{tab:dataset_mapping}, we provide a detailed mapping between the abbreviations and their corresponding full names for clarity and ease of reference. 

\section{More Experiment Results}
\label{sec:MoreExperimentResults}
\subsection{Accuracy on Long Context Tasks}
\label{sec:MoreLongBenchResults}
The detail {\it LongBench} results of LLaMA2-7b-chat and LLaMA3-8b-instruct are depicted in Table~\ref{tab1:longbench2} and Table~\ref{tab1:longbench3}. Overall, SpindleKV outperforms baselines cross various reserve ratio of KV cache. 

\subsection{LongBench Results On More Baselines}
\label{sec:MoreBaselines}
We also compared SpindleKV with three additional KV cache compression techniques on LongBench: H2O, SnapKV, and StreamingLLM. The average scores across 16 datasets for LLaMA3-8B-Instruct are showed in Table~\ref{tab:MoreBaselines}, which demonstrate the superiority of SpindleKV as a promising solution for KV cache compression compared to other method.

\begin{flushright}
    \begin{minipage}[t]{0.48\textwidth} 
        \vspace{0pt} 
        \includegraphics[width=\linewidth]{image/cos_sim_count_key.pdf}
        \captionof{figure}{Cosine similarity in Key.($\theta=0.9$)}
        \label{fig:CosineSimilarityKey}
        \vspace{1mm}
        \includegraphics[width=\linewidth]{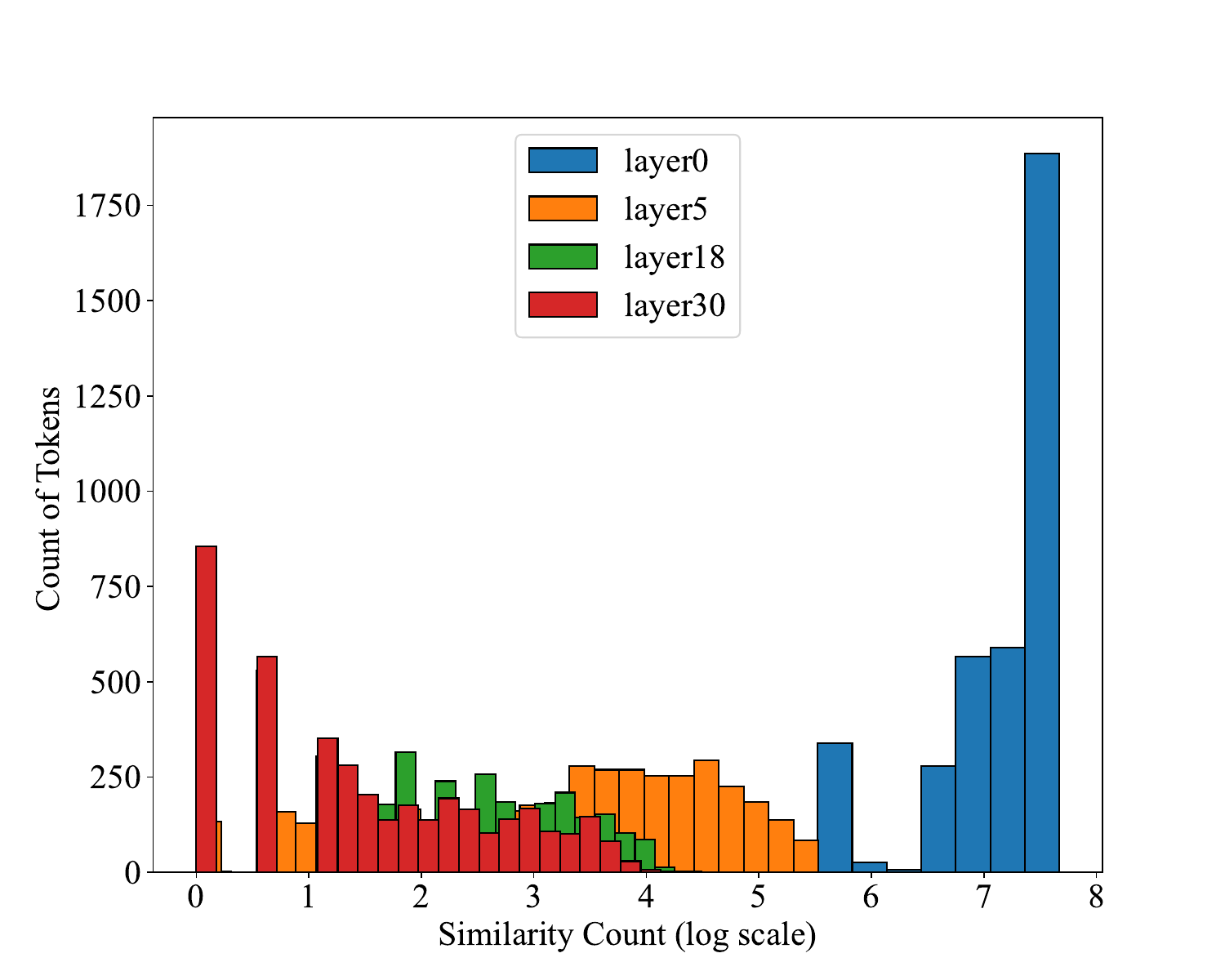}
        \captionof{figure}{Cosine similarity in Value.($\theta=0.6$)}
        \label{fig:CosineSimilarityValue}
    \end{minipage}
\end{flushright}

\begin{table}[h]
    \centering
    \renewcommand{\arraystretch}{1} %
    \begin{tabular}{ll}
        \toprule
        \textbf{Abbreviation} & \textbf{Full Name} \\
        \midrule
        Na.QA & narrativeqa \\
        Qasp  & qasper \\
        Mu.QA & multifieldqa\_en \\
        Ho.QA & hotpotqa \\
        Wi.QA & 2wikimqa \\
        Musq  & musique \\
        Gv.Rp & gov\_report \\
        QMSm  & qmsum \\
        M.New & multi\_news \\
        TREC  & trec \\
        Tr.QA & triviaqa \\
        SASm  & samsum \\
        PCnt  & passage\_count \\
        Pa.Rt & passage\_retrieval\_en \\
        Lcc   & lcc \\
        RB.P  & repobench-p \\
        \bottomrule
    \end{tabular}
    \caption{Abbreviation to Dataset Mapping}
    \label{tab:dataset_mapping}
\end{table}

\begin{table*}[h]
    \centering
    \renewcommand{\arraystretch}{1.1}
    \scalebox{0.9}{
        \begin{tabular}{l|cccc}
            \toprule
            \textbf{Reserve Ratio} &
            \textbf{H2O} &
            \textbf{SnapKV} &
            \textbf{StreamingLLM} &
            \textbf{SpindleKV} \\
            \midrule
            20.0\% & 36.12 & 39.56 & 37.46 & 40.10 \\
            40.1\% & 37.34 & 40.48 & 39.29 & 41.13 \\
            \bottomrule
        \end{tabular}
    }
    \caption{Accuracy of {\it LongBench} on LLaMA3-8b-instruct with more baselines.}
    \label{tab:MoreBaselines}
\end{table*}

\begin{table*}[h!]
    \centering
    \renewcommand{\arraystretch}{1.1}
    \scalebox{0.9}{
        \begin{tabular}{l|cc}
            \toprule
            \textbf{Model} &
            \textbf{FullKV} &
            \textbf{SpindleKV with 40\% Cache} \\
            \midrule
            LLaMA3-8B-Instruct & 22.16 token/s & 18.39 token/s \\
            Mistral-7B         & 22.48 token/s & 18.47 token/s \\
            \bottomrule
        \end{tabular}
    }
    \caption{Decoding speed (token/s) comparison between FullKV and SpindleKV with 40\% cache.}
    \label{tab:inferencespeed}
\end{table*}


\subsection{Inference Speed}
\label{sec:InferenceSpeed}
We measured the latency on LLaMA3-8B-Instruct and Mistral-7B using a context length of 4096 and a generation length of 1000 on a signal 3090 GPU. The inference speed (token/s) are showed in Table~\ref{tab:inferencespeed}, our method does not introduce significant additional time overhead during inference, which is consistent with the analysis presented in our paper.


\begin{table*}[t]
\centering
\renewcommand{\arraystretch}{1.6} %
\resizebox{\textwidth}{!}{
\begin{tabular}{lcccccccccccccccccc}
\toprule
\multirow{2}{*}{\textbf{Methods}} & \multirow{2}{*}{\textbf{Ratio}}&  \multicolumn{3}{c}{\textbf{Single-Document QA}} & \multicolumn{3}{c}{\textbf{Multi-Document QA}} & \multicolumn{3}{c}{\textbf{Summarization}} & \multicolumn{3}{c}{\textbf{Few-shot Learning}} & \multicolumn{2}{c}{\textbf{Synthetic}} & \multicolumn{2}{c}{\textbf{Code}} & \multirow{2}{*}{\textbf{AVG.}} \\
\cmidrule(lr){3-5} \cmidrule(lr){6-8} \cmidrule(lr){9-11} \cmidrule(lr){12-14} \cmidrule(lr){15-16} \cmidrule(rl){17-18}
 & & Na.QA & Qasp & Mu.QA & Ho.QA & Wi.QA & Musq & Gv.Rp & QMSm & M.New & TREC & Tr.QA & SASm & PCnt & Pa.Rt & Lcc & RB.P & \\
\midrule

FullKV &100\% & 24.50 & 31.50 & 39.36 & 43.70 & 36.23 & 21.60 & 28.50 & 23.40 & 26.39 & 74.00 & 90.48 & 42.82 & 4.77 & 69.50 & 59.21 & 54.02 & 41.87\\ 

\midrule

PyramidInfer &39.3\% &19.18 & \textbf{26.47} & 22.67 & 38.92 & 27.64 & 18.25 & 25.02 & 21.52 & 24.05 & 66.50 & 90.38 & 40.68 & 1.58 & 49.50 & 48.88 & 48.78 & 35.62 \\ 

PyramidKV &40.6\% &\textbf{24.73} & 20.75 & 35.56 & 44.00 & 32.74 & 20.77 & 23.47 & 22.62 & 22.10 & \textbf{72.00} & 90.33 & 40.61 & \textbf{5.77} & 69.25 & 58.28 & \textbf{54.85} & 39.86 \\ 

SpindleKV &39.1\% &23.87 & 26.40 & \textbf{39.02} & \textbf{44.38} & \textbf{36.02} & \textbf{22.12} & \textbf{26.10} & \textbf{23.28} & \textbf{24.24} & \textbf{72.00} & \textbf{90.43} & \textbf{41.49} & 5.23 & \textbf{69.50} & \textbf{59.37} & 54.68 & \textbf{41.13} \\ 

\midrule

PyramidInfer &31.3\% &19.58 & 23.23 & 21.60 & 36.24 & 24.45 & 16.79 & 24.31 & 21.31 & 22.78 & 62.50 & 89.74 & 40.17 & 2.20 & 49.00 & 48.04 & 49.43 & 34.46 \\ 

PyramidKV  &30.5\% &23.02 & 20.24 & 33.66 & \textbf{44.50} & 30.27 & 20.95 & 22.60 & 22.77 & 21.40 & 71.50 & 90.24 & 40.47 & \textbf{5.83} & \textbf{69.50} & 58.94 & 54.69 & 39.41 \\ 

SpindleKV &29.3\% &\textbf{24.18} & \textbf{25.71} & \textbf{37.44} & 43.42 & \textbf{34.95} & \textbf{21.97} & \textbf{25.13} & \textbf{23.52} & \textbf{23.13} & \textbf{72.00} & \textbf{90.43} & \textbf{41.47} & 5.24 & \textbf{69.50} & \textbf{59.24} & \textbf{60.04} & \textbf{41.08} \\ 

\midrule

PyramidInfer &26.0\% &18.80 & 21.61 & 17.34 & 33.47 & 22.29 & 13.95 & 23.32 & 21.08 & 22.01 & 61.00 & 87.85 & 40.28 & 2.25 & 32.50 & 49.57 & 50.83 & 32.38 \\ 

PyramidKV &24.2\% &24.13 & 19.70 & 33.08 & 43.32 & 30.86 & 21.08 & 22.16 & 23.02 & 20.59 & \textbf{72.00} & 90.16 & 40.00 & 5.37 & \textbf{69.50} & 58.48 & 53.92 & 39.21 \\ 

SpindleKV &23.6\% &\textbf{24.66} & \textbf{23.73} & \textbf{34.65} & \textbf{43.54} & \textbf{33.74} & \textbf{22.15} & \textbf{24.44} & \textbf{23.11} & \textbf{22.92} & 71.50 & \textbf{90.56} & \textbf{41.38} & \textbf{5.58} & \textbf{69.50} & \textbf{58.57} & \textbf{54.13} & \textbf{40.26} \\ 

\midrule

PyramidInfer &21.7\%  &16.98 & 15.74 & 17.47 & 31.30 & 22.74 & 14.68 & 23.09 & 20.72 & 21.31 & 54.50 & 84.45 & 40.32 & 2.36 & 21.00 & 53.22 & 51.01 & 30.68 \\ 

PyramidKV &21.6\% &23.77 & 18.77 & 34.46 & 42.84 & 30.46 & 21.00 & 22.19 & \textbf{22.98} & 20.23 & \textbf{72.50} & 90.18 & 40.05 & \textbf{5.70} & \textbf{69.50} & 57.33 & 53.83 & 39.11 \\ 

SpindleKV &21.2\%  &\textbf{23.92} & \textbf{23.16} & \textbf{35.87} & \textbf{43.52} & \textbf{33.90} & \textbf{21.20} & \textbf{24.27} & 22.66 & \textbf{22.46} & 71.50 & \textbf{90.56} & \textbf{41.37} & 5.58 & \textbf{69.50} & \textbf{58.26} & \textbf{53.93} & \textbf{40.10} \\ 

\midrule

PyramidInfer &16.6\% &15.60 & 16.31 & 15.89 & 30.04 & 20.58 & 10.43 & 22.55 & 20.03 & 21.10 & 52.00 & 78.46 & 39.50 & 1.30 & 13.03 & 56.74 & 51.28 & 29.05 \\ 

PyramidKV &16.1\% &22.73 & 17.58 & 34.83 & 43.86 & 27.50 & \textbf{21.66} & 21.44 & 22.47 & 19.28 & 71.00 & 88.93 & 39.88 & \textbf{5.59} & \textbf{69.50} & 56.62 & 53.39 & 38.51 \\ 

SpindleKV &16.0\% &\textbf{24.34} & \textbf{20.99} & \textbf{35.72} & \textbf{44.06} & \textbf{31.29} & 20.52 & \textbf{23.22} & \textbf{22.79} & \textbf{21.90} & \textbf{71.50} & \textbf{90.33} & \textbf{40.60} & 5.52 & \textbf{69.50} & \textbf{57.94} & \textbf{53.93} & \textbf{39.63} \\ 
\bottomrule
\end{tabular}
}
\caption{LongBench Results of LLaMA3-8b-instruct. }
\label{tab1:longbench2}
\vspace{-3mm}
\end{table*}

\begin{table*}[t]
\centering
\renewcommand{\arraystretch}{1.6} %
\resizebox{\textwidth}{!}{
\begin{tabular}{lcccccccccccccccccc}
\toprule
\multirow{2}{*}{\textbf{Methods}}& \multirow{2}{*}{\textbf{Ratio}} & \multicolumn{3}{c}{\textbf{Single-Document QA}} & \multicolumn{3}{c}{\textbf{Multi-Document QA}} & \multicolumn{3}{c}{\textbf{Summarization}} & \multicolumn{3}{c}{\textbf{Few-shot Learning}} & \multicolumn{2}{c}{\textbf{Synthetic}} & \multicolumn{2}{c}{\textbf{Code}} & \multirow{2}{*}{\textbf{AVG.}} \\
\cmidrule(lr){3-5} \cmidrule(lr){6-8} \cmidrule(lr){9-11} \cmidrule(lr){12-14} \cmidrule(lr){15-16} \cmidrule(rl){17-18}
 && Na.QA & Qasp & Mu.QA & Ho.QA & Wi.QA & Musq & Gv.Rp & QMSm & M.New & TREC & Tr.QA & SASm & PCnt & Pa.Rt & Lcc & RB.P & \\
\midrule

FullKV &100\% & 18.39 & 20.14 & 35.67 & 30.92 & 25.73 & 10.64 & 25.58 & 20.98 & 26.43 & 64.00 & 83.38 & 41.02 & 5.50 & 10.00 & 60.81 & 55.12 & 33.39
\\ 

\midrule

PyramidInfer &40.7\% & 15.36 & 15.40 & 19.23 & 29.14 & 24.53 & 7.49 & 21.64 & 19.66 & 22.70 & 54.00 & 81.79 & 40.71 & 4.00 & 3.50 & 54.29 & 51.98 & 29.09 \\
PyramidKV &41.3\% & 18.38 & 20.99 & 35.98 & 30.76 & 25.45 & \textbf{10.79} & 23.73 & \textbf{20.88} & 25.08 & \textbf{64.00} & 83.75 & \textbf{41.17} & \textbf{6.00} & \textbf{10.50} & \textbf{60.58} & \textbf{54.93} & 33.31 \\
SpindleKV &41.1\% &  \textbf{18.45} & \textbf{21.23} & \textbf{36.67} & \textbf{30.80} & \textbf{25.74} & 10.62 & \textbf{24.49} & 20.66 & \textbf{25.18} & \textbf{64.00} & \textbf{84.31} & 41.11 & \textbf{6.00} & 10.00 & 60.45 & 54.92 & \textbf{33.41} \\

\midrule

PyramidInfer &31.2\% &13.80 & 15.27 & 17.69 & 27.69 & \textbf{26.10} & 7.27 & 20.53 & 19.42 & 22.04 & 53.50 & 77.06 & 40.50 & 2.00 & 6.50 & 52.69 & 50.46 & 28.28 \\
PyramidKV &30.8\% &17.78 & \textbf{20.49} & 36.86 & 30.55 & 26.04 & \textbf{9.93} & 22.92 & \textbf{20.97} & 24.13 & \textbf{64.00} & 83.59 & 41.06 & \textbf{6.00} & \textbf{11.00} & \textbf{60.74} & \textbf{54.15} & \textbf{33.13} \\
SpindleKV &30.8\% &\textbf{18.13} & 19.95 & \textbf{36.91} & \textbf{30.77} & \textbf{26.10} & 9.91 & \textbf{23.19} & 20.95 & \textbf{24.35} & \textbf{64.00} & \textbf{83.69} & \textbf{41.37} & \textbf{6.00} & 9.00 & 59.85 & 53.77 & 33.00
\\ 

 
\midrule
PyramidInfer &25.9\% &14.85 & 15.19 & 15.41 & 26.82 & 24.95 & 5.67 & 19.96 & 18.66 & 21.34 & 49.00 & 76.22 & 38.83 & 4.50 & 5.00 & 51.87 & 51.71 & 27.50 \\
PyramidKV &26.3\% &\textbf{17.32} & \textbf{20.99} & 36.37 & 30.86 & 25.62 & 9.80 & 22.38 & 20.57 & 23.24 & \textbf{64.00} & 83.81 & 40.82 & \textbf{6.00} & \textbf{10.50} & 59.95 & \textbf{54.40} & 32.91 \\
SpindleKV &26.0\% &17.31 & 20.72 & \textbf{36.86} & \textbf{31.09} & \textbf{25.80} & \textbf{10.00} & \textbf{22.93} & \textbf{21.32} & \textbf{23.94} & \textbf{64.00} & \textbf{84.00} & \textbf{40.97} & \textbf{6.00} & 10.00 & \textbf{60.92} & 53.70 & \textbf{33.10}
\\ 

\midrule

PyramidInfer &21.10\% &13.39 & 14.44 & 13.23 & \textbf{30.97} & \textbf{27.24} & 8.27 & 19.67 & 18.71 & 20.49 & 43.50 & 70.34 & 37.94 & 3.00 & 2.00 & 52.44 & 51.41 & 26.69 \\
PyramidKV &22.3\% &\textbf{17.77} & \textbf{21.77} & 35.70 & 30.78 & 25.96 & \textbf{9.99} & 21.74 & 20.50 & \textbf{23.48} & \textbf{64.00} & 83.80 & 40.33 & \textbf{6.00} & \textbf{10.00} & 59.69 & 53.83 & 32.83 \\
SpindleKV &22.1\% &17.34 & 21.20 & \textbf{35.81} & 30.69 & 25.64 & 9.90 & \textbf{22.31} & \textbf{20.64 }& 23.17 & \textbf{64.00} & \textbf{84.15} & \textbf{40.65} & \textbf{6.00} & \textbf{10.00} & \textbf{60.31} & \textbf{54.49} & \textbf{32.89}
\\ 

\midrule

PyramidInfer &15.9\% &12.11 & 14.49 & 14.25 & 26.98 & \textbf{27.30} & 6.76 & 19.28 & 18.33 & 19.78 & 38.00 & 61.76 & 38.84 & 2.00 & 6.00 & 52.35 & 50.78 & 25.56 \\
PyramidKV &16.8\% &16.95 & \textbf{20.81} & \textbf{36.05} & \textbf{31.22} & 25.50 & 9.69 & 20.77 & \textbf{20.52} & 22.53 & \textbf{64.00} & 83.72 & \textbf{40.09} & \textbf{6.00} & 10.00 & 58.36 & 53.33 & 32.47 \\
SpindleKV &16.7\% &\textbf{17.33} & 20.64 & 35.04 & 31.01 & 25.88 & \textbf{9.72} & \textbf{21.28} & 20.15 & \textbf{22.67} & \textbf{64.00} & \textbf{87.84} & 39.81 & 5.50 & \textbf{11.00} & \textbf{59.60} & \textbf{54.34} & \textbf{32.86} \\

\bottomrule
\end{tabular}
}
\caption{LongBench Results of LLaMA2-7b-chat. }
\label{tab1:longbench3}
\vspace{-3mm}
\end{table*}

\end{document}